\pdfoutput=1

\documentclass[11pt]{article}

\usepackage{ACL2023}

\usepackage{times}
\usepackage{latexsym}

\usepackage[T1]{fontenc}

\usepackage[utf8]{inputenc}

\usepackage{microtype}

\usepackage{inconsolata}

\usepackage{multirow}
\usepackage{tabularx}
\usepackage{booktabs}
\usepackage{graphicx}

\usepackage{enumitem}
\usepackage{booktabs}
\usepackage{array}
\usepackage{hyperref}
\usepackage{url}
\usepackage{multirow}
\usepackage{colortbl}
\usepackage{booktabs}
\usepackage{amssymb}
\usepackage{amsmath}
\usepackage{amsthm}
\usepackage{graphicx}
\usepackage{makecell}
\usepackage{threeparttable}
\usepackage{CJKutf8}
\usepackage{lscape}
\usepackage{fancyhdr}
\usepackage[ruled,vlined]{algorithm2e}

\usepackage{threeparttable}

\usepackage{caption}
\usepackage{graphicx}
\usepackage{float} 
\usepackage{subfigure}
\usepackage{soul}

\title{Farewell to Aimless Large-scale Pretraining: Influential Subset Selection for Language Model}

\author{
    {\normalsize
     \textbf{Xiao Wang}$^{\bigstar*}$, 
     \ \ Weikang Zhou$^{\bigstar}$\thanks{$^*$  Equal contribution.} ,
     \ \ \textbf{Qi Zhang}$^{\bigstar}$$^{\dagger}$, 
     \ \ Jie Zhou$^{\bigstar}$, 
     \ \ Songyang Gao$^{\bigstar}$,}\\
    {\normalsize
     \textbf{Junzhe Wang}$^{\bigstar}$, 
    \ \ \textbf{Menghan Zhang}$^{\blacklozenge}$, 
    \ \ \textbf{Xiang Gao}$^{\clubsuit}$, 
    \ \ \textbf{Yunwen Chen}$^{\clubsuit}$,
    \ \ \textbf{Tao Gui}$^{\blacklozenge}$
    \thanks{{} {} Corresponding Author} 
    }\\
  {$^\bigstar$ \normalsize School of Computer Science, Fudan University, Shanghai, China} \\
  {$^\blacklozenge$ \normalsize Institute of Modern Languages and Linguistics, Fudan University, Shanghai, China} \\
  {$^\clubsuit$ \normalsize DataGrand Information Technology (Shanghai) Co., Ltd.} \\
  \texttt{\normalsize \{xiao\_wang20,qz,tgui\}@fudan.edu.cn}
}

\begin{document}
\maketitle
\begin{abstract}
Pretrained language models have achieved remarkable success in various natural language processing tasks.
However, pretraining has recently shifted toward larger models and larger data, and this has resulted in significant computational and energy costs.
In this paper, we propose Influence Subset Selection (ISS) for language model, which explicitly utilizes end-task knowledge to select a tiny subset of the pretraining corpus.
Specifically, the ISS selects the samples that will provide the most positive influence on the performance of the end-task. Furthermore, we design a gradient matching based influence estimation method, which can drastically reduce the computation time of influence.
With only 0.45\% of the data and a three-orders-of-magnitude lower computational cost, ISS outperformed pretrained models (e.g., RoBERTa) on eight datasets covering four domains.

\end{abstract}

\section{Introduction}



Pretrained language models (PTMs) \cite{peters-etal-2018-deep, devlin-etal-2019-bert, liu2019roberta}, trained on massive and heterogeneous corpora, have significantly improved the state-of-the-art across a variety of natural language processing tasks \cite{wang-etal-2022-miner, wang2023instructuie}.
\citet{kaplan2020scaling} found power laws relating cross entropy loss to the sizes of language models and their training datasets. 
As a result, the field has recently shifted toward larger models and large data \cite{brown2020language, rae2021scaling, smith2022using, chowdhery2022palm} in hopes of improving performance.

However, training a state-of-the-art language model requires substantial computational resources which demand considerable energy, along with the associated financial and environmental costs \cite{strubell-etal-2019-energy}.
For example, RoBERTa-Large \cite{liu2019roberta}, which was trained on 1000 V100 GPUs for approximately one day, has a computational cost of $4.36 \times 10^{21}$ FLOPs. Recently, \citet{chowdhery2022palm} proposes PaLM, which consumes 580 times more FLOPs than RoBERTa-Large. PaLM was trained on 6144 TPU v4 chips for more than 1200 hours, which is unaffordable for most researchers. 
Therefore, finding ways to speed up pretraining is crucial for the development of pretrained model research.





\begin{table}[t]
\centering
\small
\renewcommand\arraystretch{1.2}
\setlength{\tabcolsep}{2.8pt}
\begin{threeparttable}
 \resizebox{1\columnwidth}{!}{
\begin{tabular}{l|l|l|l}
\hline
& \textbf{PLMs}  & \textbf{TLM} & \textbf{ISS} \\ 
\hline
Training Data & The entire $\mathcal{D}$                                & \begin{tabular}[c]{@{}l@{}}Subset of $\mathcal{D}$ \\ \& task data $\mathcal{T}$\end{tabular} 
& \begin{tabular}[c]{@{}l@{}}Subset of $\mathcal{D}$\\ \& task data $\mathcal{T}$\end{tabular} \\ \hline
Compute Cost  & \begin{tabular}[c]{@{}l@{}}240000\\ GPU·hours \end{tabular} & \begin{tabular}[c]{@{}l@{}}240 \\GPU·hours\end{tabular} & \begin{tabular}[c]{@{}l@{}}80 \\GPU·hours\end{tabular} \\ \hline
Generality    & Task-Agnostic  & $X$-Dep  & $X \& Y$-Dep        \\ \hline
\end{tabular}
}
\end{threeparttable}

\caption{Qualitative comparison between PLMs, TLM, and ISS(ours). 
X/Y-Dep means the pretraining data is X/Y dependent.}
\label{table:comparison}
\vspace{-4mm}
\end{table}


In general, there are three main strategies used to speed up pretraining in NLP: parallel architectures, efficient model architectures, and novel pretraining tasks. 
The first one is to train a single model utilizing multiple GPUs distributed in many computational nodes \cite{wang2020minilm, shazeer2018mesh, huang2019gpipe}. Unfortunately, the gains in efficiency of this strategy depend entirely on the amount of computing hardware used. 
The second strategy is to improve model structures to reduce the computational complexity and therefore improve efficiency \cite{wang2020linformer,katharopoulos2020transformers,roy2021efficient}.
The last one explores more challenging pretraining tasks to accelerate a model's convergence \cite{clark2019electra, joshi2020spanbert, levine2020pmi}.
However, their improvements are limited, with a reduction of less than an order of magnitude in computational expenses (measured in FLOPs).

In this paper, we aim to reduce the computational costs from data level (See Table \ref{table:comparison}). 
The PLMs are trained on the entire pretraining corpus $D$, which is task-agnostic. 
To take the downstream task into account, we hope to select the most relevant samples from the pretraining corpus based on the downstream data. 
Recently, \citet{yao2022nlp} proposes TLM, which retrieves data from a pretraining corpus using task data as queries. However, TLM remains task-agnostic, because it only considers text (i.e., X) similarities and ignores the label (i.e., Y) information.

Motivated by influence function \cite{cook1982residuals,koh2017understanding}, we propose Influential Subset Selection (ISS) for language model, i.e. selecting the samples with the most positive influence on the downstream task.
To calculate the label-aware influence value, ISS utilizes the derivation chain rule from a test objective to training samples. 
Nevertheless, directly applying the chain rule leads to computing the inverse of Hessian with the complexity of $O(nq^2 + q^3)$(n is the number of examples and q is parameter size), which is computationally expensive and may run out-of-memory in neural networks.
To address this problem, we propose a gradient matching based influence approximation method for selecting pretraining data, which estimates the influence score by matching the gradient values of pretraining samples and end-task samples. 
Our method avoids the computation of the inverse of Hessian and significantly speeds up the estimation time of influence.


Our main contributions are summarized as follows:

\begin{itemize}[leftmargin=*, align=left]
    \item We propose Influential Subset Selection for language model, which explicitly utilizes knowledge of the end-task to select the pretraining corpus.
    \item We design a simple, efficient, gradient matching based method for influence estimation, which avoids the calculation of the inverse of Hessian and significantly speeds up the estimation time. 
    \item We evaluate the effectiveness of our method on eight tasks covering four domains. Notably, ISS outperforms PTMs (e.g. RoBERTa) with only \textbf{0.45\% of the data} and \textbf{three orders of magnitude reduced FLOPS}. Our code can be found at \url{https://github.com/nitwtog/ISS}.

\end{itemize}

\section{Preliminaries}
\subsection{Definition}
We assume an end-task dataset represented as $\mathcal{T} = \left(\mathcal{Z}_t\right)$ where $\mathcal{Z}_t=\left\{(x_t^1, y_t^1), (x_t^2, y_t^2), \ldots, (x_t^m, y_t^m)\right\}$ represents a set of texts with their ground truth labels. And we assume a large-scale pretraining corpus $\mathcal{D} = \left(\mathcal{Z}_p\right)$, where $\mathcal{Z}_p=\left\{x_p^1, x_p^2, \ldots, x_p^M \right\}$ represents unlabeled data. 
We define $f=f^{(\text {head})} \circ f^{(\text {feat})}$, such that $f^{(\text {feat})}(\cdot; \theta \in \Theta)$ is a feature extractor that is transferable across learning stages (e.g. pretraining to finetuning) and $f^{(\text {head})}(\cdot; \phi \in \Phi)$ is a task-specific head that is not transferable. And we assume $l_p({z_p, \theta, \phi_p})$ and  $l_t({z_t, \theta, \phi_t})$ are the loss functions of pretraining and end-task. 

\subsection{Influence Function}
Influence function \cite{cook1982residuals,koh2017understanding} provides an efficient way to estimate the importance of a training sample.
Considering a training sample $z$ was weighted by a small $\epsilon$ during training, 
the empirical risk minimizer can be written as 
\begin{equation}
\hat{\theta}_{\epsilon, z}=\arg \min _{\theta \in \Theta} \frac{1}{n} \sum_{z_i \in \mathcal{D}} l\left(z_i, \theta\right)+\epsilon \cdot l(z, \theta)
\end{equation} 
Assigning $- \frac{1}{n}$ to $\epsilon$ is equivalent to removing the training example $z_p$. 
Then, the influence of weighting $z_p$ on the parameters is given by
\begin{equation}
\mathcal{I}_{\text {param }}(z)=\left.\frac{\mathrm{d} \hat{\theta}_{\epsilon, z}}{\mathrm{~d} \epsilon}\right|_{\epsilon=0}=-H_{\hat{\theta}}^{-1} \nabla_\theta l(z, \hat{\theta})
\end{equation}
where $H_{\hat{\theta}}=\frac{1}{n} \sum_{z_i \in \mathcal{D}} \nabla_\theta^2 l\left(z_i, \hat{\theta}\right)$ is the Hessian and positive definite by assumption, $\mathcal{I}_{\text {param }}(z) \in R^N$ , N is the number of network parameters.
Then, we can linearly approximate the parameter change due to removing z without retraining the model by computing $\hat{\theta}_{- z} - \hat{\theta} \approx - \frac{1}{n} \mathcal{I}_{\text {param }}(z)$.

\section{Methodology}
We investigate an influence-based subset selection method to perform efficient pretraining while attempting to minimize accuracy loss on the end-task dataset (Section \ref{sect:label-dependent influence estimation}). 
Due to the high computational costs of influence function \cite{koh2017understanding}, we design an influence approximation strategy to speed up the calculation (Section \ref{sect:influence approximation}).

\subsection{Influence of Pretraining Corpus}
\label{sect:label-dependent influence estimation}
PTMs used in previous works usually adopt language modeling as pretraining tasks, lacking task-specific prior knowledge.
However, we often know the end-task beforehand, so we can make specific choices about our pretraining regimen to improve end-task performance.
Under this setting, we introduce Influential Subset Selection for language model, which measures the importance of pretraining samples by considering the X and Y information of the end-task simultaneously.

Specifically, pretraining sample $z_p$ affects the prediction of end-task sample $z_t$ by influencing the parameters of the feature encoder $\theta$. We can apply the chain rule to measure the influence of upweighting pretraining sample $z_p$ on the loss at end-task sample $z_t$.
\begin{equation}
\begin{aligned}
\mathcal{I}\left(z_p, z_t\right) & \left.\triangleq \frac{d l\left(z_{t}, \hat{\theta}_{\epsilon, z}\right)}{d \epsilon}\right|_{\epsilon=0} \\
& =\left.\nabla_\theta l\left(z_{t}, \hat{\theta}\right)^{\top} \frac{d \hat{\theta}_{\epsilon, z}}{d \epsilon}\right|_{\epsilon=0} \\
& =-\nabla_\theta l\left(z_{t}, \hat{\theta}\right)^{\top} H_{\hat{\theta}}^{-1} \nabla_\theta l(z_p, \hat{\theta})
\end{aligned}
\end{equation}

The more negative $\mathcal{I}\left(z_p, z_t\right)$ is, the more positive influence $z_p$ can provide.
However, computing the Hessian for the full training dataset is expensive, and inverting it is similarly prohibitive: with $n$ training data points and $p$ parameters, this computation requires $O(n*p^2 +p^3)$ operations. It means that evaluating the influence of large-scale pretrained corpus is not achievable. 
Thus, we propose an influence approximation algorithm to speed up the estimation time.

\subsection{Influence Approximation}
\label{sect:influence approximation}
Motivated by calculus, the update of the model parameters is the result of cumulative updates over several training iterations.
Similarly, the difference between the loss of test point $z_t$ at the end of training versus at the beginning of training can be decomposed along the path taken by the training process.
Thus, we hypothesize that the influences of all training examples on a fixed test point $z_t$ is exactly the total reduction in loss on $z_t$.

Assume that we train the feature encoder by minimizing the pertaining loss $l_{p}(z_p;\theta, \phi)$, via an iterative optimization procedure (such as SGD) which utilizes one training example $z_p$ in iteration t. 
The parameters of the feature encoder before and after iteration t are $\theta_t$ and $\theta_{t+1}$ respectively.
The influence of $z_t$ on $z_p$ can be approximated in the following way.
\begin{equation}
\mathcal{I}\left(z_p, z_t\right) = l_{t}\left( z_p, \theta_t\right)-l_{t}\left(z_p, \theta_{t+1}\right)
\end{equation}
Suppose we are at point $\theta_t$, and
we make a first-order Taylor expansion of function $l_{p}\left(z_p, \theta_{t+1}\right)$.

\begin{equation}
\resizebox{1\columnwidth}{!}{
$\begin{aligned}
\small l_{t}\left(z_p, \theta_{t+1}\right)=&
l_{t}\left(z_p,\theta_t\right)+\nabla_\theta l_{t}\left(z_p,\theta_t\right) \cdot\left(\theta_{t+1}-\theta_t\right)\\
&+O\left(\left\|\theta_{t+1}-\theta_t\right\|^2\right)
\end{aligned}$
}
\label{Taylor}
\end{equation}

\begin{figure}[t]
\centering
  \includegraphics[width=3.0in]{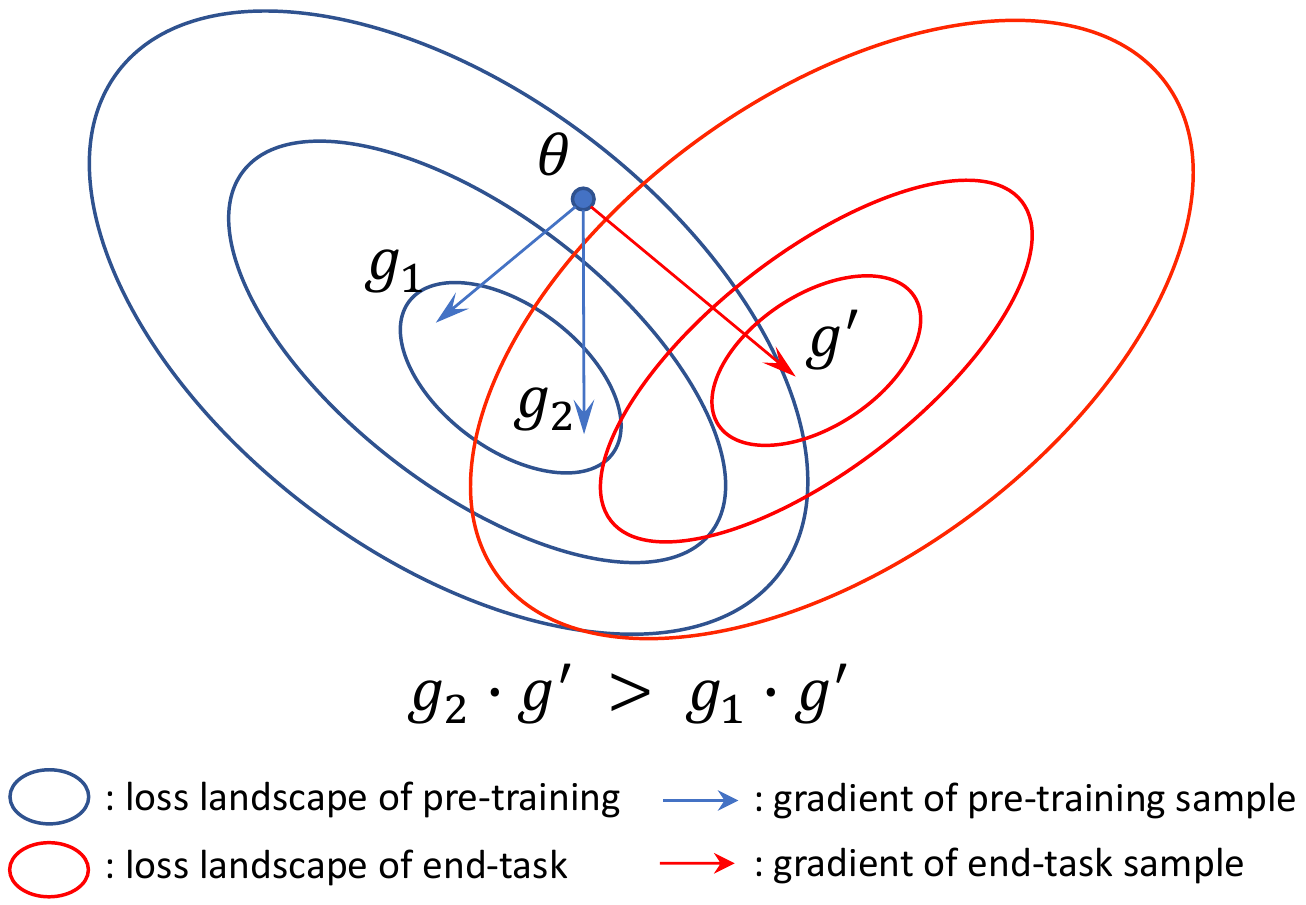}
  \caption{Illustration of gradient matching based influence approximation. $g_1$ and $g_2$ are the loss gradients of two different pretrained samples respectively, while $g'$ is the loss gradient of the end-task sample. 
  The influence of a pretrained sample is measured by how a small step based on its gradient affects the loss on the end-task sample.
  Compared to $g_1$, the update step of $g_2$ is more generalized.
  }
 \label{GradMatch}
 \vspace{-3mm}
\end{figure}

Assuming the model employs SGD as the optimizer, then the update in parameters is $\theta_{t+1}-\theta_t=-\eta_t \nabla_\theta l_p\left(z_t,\theta_t\right)$, where $\eta_t$ is the learning rate at iteration $t$.
Eq. (\ref{Taylor}) guarantees approximation precision as long as the update magnitude of $\theta$ is sufficiently small.
By substituting the parameter update formula and disregarding the higher-order term, we arrive at the following first-order approximation.
\begin{equation}
\small l_{t}\left(z^{\prime}, \theta_t\right)-l_{t}\left( z^{\prime}, \theta_{t+1}\right) \approx \eta_t \nabla_\theta l_{t}\left(z^{\prime}, \theta_t\right) \cdot \nabla_\theta l_{p}\left(z_t, \theta_t\right)
\end{equation}
We refer to this first-order approximation as gradient matching-based influence estimation. 
The full algorithm is provided in Algorithm \ref{GradMatch}.

\textbf{Visualisation}  We visualize our influence estimation method in Fig \ref{GradMatch}. 
$g_1$ and $g_2$ are the loss gradients of two different pretrained samples respectively, while $g'$ is the loss gradient of the end-task sample.
The influence of a pretrained sample can be viewed as the dot product of its gradient and the gradient of the end-task sample.
Higher influence suggests that a network is learning parameters that generalize.


\begin{algorithm}[t]
\SetKwInput{KwRequire}{Require}

\DontPrintSemicolon
\SetAlgoLined
\KwRequire{Pretraining corpus $\mathcal{D}$;
task training set  $\mathcal{T}_t$ and validation set  $\mathcal{T}_v$;
learning rate $\alpha$;
initial subset $S$;
candidates size k.
}

Random initialize network $\theta, \phi_p, \phi_t$

\wx$\hat{\theta}, \hat{\phi_p}, \hat{\phi_t} = \arg \min \frac{1}{n} \sum_{z_i \in \mathcal{T}_t} l_{p}\left(z_i\right)+ l_{t}(z_i)$

\For{$z_p \in \mathcal{D}$}{
 Compute $\nabla_\theta l_{p}\left(z_p, \hat{\theta}, \hat{\phi_p}\right) $
}

\For{$z^{\prime} \in \mathcal{T}_v$}{
 Compute $\nabla_\theta l_{t}\left(z^{\prime}, \hat{\theta}, \hat{\phi_t}, \right)$ 
 
     \For{$z_p \in \mathcal{D}$}{
     $\mathcal{I}\left(z_p, z^{\prime}\right)= \nabla_\theta l_{p}\left(z_p, \hat{\theta}, \hat{\phi_p}\right) \cdot 
     \nabla_\theta l_{t}\left(z^{\prime}, \hat{\theta}, \hat{\phi_t}, \right)$
    }
    Sort pretraining samples based on influence
    
    Add top k influential samples to $S$
}

Return influential subset $S$
\caption{Influential Subset Selection for Language Model\label{IR}}
\label{ISS}
\end{algorithm}

\subsection{Implementation Details}
\label{implementDetails}
Based on the influence score, we select the most relevant samples from the pretraining corpus. 
Following TLM, we first select a subset via a BM25 retrieval method. 
Then, we compute the influence score based on this subset to make ISS scalable and efficient.

Moreover, the number of parameters in large-scale language models is very large, leading to very high dimensional gradients. 
To tackle this problem, we adopt a last-layer gradient approximation by only considering the last layer gradients of pretrained encoder. 
We select a subset of mini-batches by matching the weighted sum of mini-batch pretraining gradients to the mini-batch task gradients. 
Let $B_p$ and $B_t$ be the batch size of pretraining and end-task. 
The use of mini-batches considerably reduces the number of selection rounds during the ISS algorithm by a factor of B, resulting in $B_p*B_t$ speed up.

\section{Experimental Setup}
To evaluate the efficiency and generality of our approach, we conduct experiments in two settings: pretraining from scratch, and further pretraining.

\subsection{Pretraining from Scratch}
\begin{table}[b]
\vspace{-5mm}
\small
\centering
\renewcommand\arraystretch{1.2}
\setlength{\tabcolsep}{3.0pt}
 \resizebox{1\columnwidth}{!}{
\begin{tabular}{llrrrr}
\hline \textbf{Domain} & \textbf{Task} & \textbf{Train} & \textbf{Dev.} & \textbf{Test} & \textbf{Classes} \\
\hline \multirow{2}{*}{ BIOMED } & CHEMPROT & 4169 & 2427 & 3469 & 13 \\
& ${ }^{\dagger}$ RCT & 18040 & 30212 & 30135 & 5 \\
\hline \multirow{2}{*}{ CS } & ACL-ARC & 1688 & 114 & 139 & 6 \\
& SCIERC & 3219  & 455 & 974 & 7 \\
\hline \multirow{2}{*}{ NEWS } & HYPERPARTISAN & 515 & 65 & 65 & 2 \\
& ${ }^{\dagger}$ AGNEWS & 115000 & 5000 & 7600 & 4 \\
\hline \multirow{2}{*}{ REVIEWS } & ${ }^{\dagger}$ HELPFULNESS & 115251 & 5000 & 25000 & 2 \\
& ${ }^{\dagger}$ IMDB & 20000  & 5000 & 25000 & 2 \\
\hline
\end{tabular}
}
\caption{Statistics of various target datasets.
${ }^{\dagger}$ indicates high-resource settings.}
\label{table:datasets}
\end{table}

\begin{table*}[ht]
\centering
\renewcommand\arraystretch{1.1}
\setlength{\tabcolsep}{2pt}
\resizebox{\textwidth}{65mm}
{
 \begin{threeparttable}

\begin{tabular}{l|ccc|cccccccccllll}
\cline{1-13}
\textbf{Model} & \textbf{Param} & \textbf{Data}\tnote{1} & \textbf{FLOPs}\tnote{2} & \textbf{AGNews}                                                & \textbf{Hyp.}                                                   & \textbf{Help.}                                                  & \textbf{IMDB}                                                  & \textbf{ACL.}                                                    & \textbf{SciERC}                                                & \textbf{Chem.}                                                   & \textbf{RCT}                                                   & \textbf{Avg.}   &                      &                               &  &  \\ \cline{1-13}
Bert-Base        & 109M             & 16G           & 2.79E19          & \begin{tabular}[c]{@{}c@{}}93.50\\ \small±0.15\end{tabular}           & \begin{tabular}[c]{@{}c@{}}91.93\\ \small±1.74\end{tabular}          & \begin{tabular}[c]{@{}c@{}}69.11\\ \small±0.17\end{tabular}          & \begin{tabular}[c]{@{}c@{}}\ul{93.77}\\ \small±0.22\end{tabular}          & \begin{tabular}[c]{@{}c@{}}69.45\\ \small±2.90\end{tabular}           & \begin{tabular}[c]{@{}c@{}}\ul{80.98}\\ \small±1.07\end{tabular}          & \begin{tabular}[c]{@{}c@{}}81.94\\ \small±0.38\end{tabular}           & \begin{tabular}[c]{@{}c@{}}87.00\\ \small±0.06\end{tabular}          & 83.46           &                      &                               &  &  \\
Bert-Large       & 355M             & 16G           & 9.07E19          & \begin{tabular}[c]{@{}c@{}}93.51\\ \small±0.40\end{tabular}          & \begin{tabular}[c]{@{}c@{}}91.62\\ \small±0.69\end{tabular}          & \begin{tabular}[c]{@{}c@{}}69.39\\ \small±1.14\end{tabular}          & \begin{tabular}[c]{@{}c@{}}\textbf{94.76}\\ \small±0.09\end{tabular} & \begin{tabular}[c]{@{}c@{}}69.13\\ \small±2.93\end{tabular}           & \begin{tabular}[c]{@{}c@{}}\textbf{81.37}\\ \small±1.35\end{tabular} & \begin{tabular}[c]{@{}c@{}}\textbf{83.64}\\ \small±0.41\end{tabular}  & \begin{tabular}[c]{@{}c@{}}\textbf{87.13}\\ \small±0.09\end{tabular} & 83.82           &                      &                               &  &  \\
TLM\scriptsize{(Small)}        & 109M             & 0.91G         & 2.74E18          & \begin{tabular}[c]{@{}c@{}}\ul{93.74}\\ \small±0.20\end{tabular}          & \begin{tabular}[c]{@{}c@{}}\ul{93.53}\\ \small±1.61\end{tabular} & \begin{tabular}[c]{@{}c@{}}\ul{70.54}\\ \small±0.39\end{tabular}          & \begin{tabular}[c]{@{}c@{}}93.08\\ \small±0.17\end{tabular}          & \begin{tabular}[c]{@{}c@{}}\ul{69.84}\\ \small±1.53\end{tabular}           & \begin{tabular}[c]{@{}c@{}}80.51\\ \small±1.53\end{tabular}          & \begin{tabular}[c]{@{}c@{}}\ul{81.99}\\ \small±0.42\end{tabular}           & \begin{tabular}[c]{@{}c@{}}\ul{86.99}\\ \small±0.03\end{tabular}          & \ul{83.78}           &                      &                               &  &  \\
TLM\scriptsize{(Small-20\%)}\tnote{3}   & 109M             & 0.18G        & 1.82E18          & \begin{tabular}[c]{@{}c@{}}93.57\\ \small±0.21\end{tabular}          & \begin{tabular}[c]{@{}c@{}}93.11\\ \small±0.46\end{tabular}          & \begin{tabular}[c]{@{}c@{}}70.02\\ \small±0.40\end{tabular}         & \begin{tabular}[c]{@{}c@{}}93.20\\ \small±0.03\end{tabular}          & \begin{tabular}[c]{@{}c@{}}67.27\\ \small±2.85\end{tabular}           & \begin{tabular}[c]{@{}c@{}}78.87\\ \small±0.63\end{tabular}          & \begin{tabular}[c]{@{}c@{}}80.80\\ \small±0.63\end{tabular}           & \begin{tabular}[c]{@{}c@{}}86.65\\ \small±0.01\end{tabular}          & 82.93          &                      &                               &  &  \\
ISS\scriptsize{(Small-scale)}  & 109M             & 0.18G        & 1.82E18          & \begin{tabular}[c]{@{}c@{}}\textbf{93.78}\\ \small±0.06\end{tabular} & \begin{tabular}[c]{@{}c@{}}\textbf{93.53}\\ \small±0.00\end{tabular}           & \begin{tabular}[c]{@{}c@{}}\textbf{70.78}\\ \small±0.29\end{tabular} & \begin{tabular}[c]{@{}c@{}}93.25\\ \small±0.07\end{tabular}          & \begin{tabular}[c]{@{}c@{}}\textbf{72.41}\\ \small±0.66\end{tabular}  & \begin{tabular}[c]{@{}c@{}}80.56\\ \small±0.43\end{tabular}          & \begin{tabular}[c]{@{}c@{}}81.71\\ \small±0.10\end{tabular}           & \begin{tabular}[c]{@{}c@{}}\ul{86.99}\\ \small±0.02\end{tabular}          & \textbf{84.11} & \multicolumn{1}{c}{} & \multicolumn{1}{c}{}          &  &  \\ \cline{1-13}
RoBERTa-Base     & 125M             & 160G          & 1.54E21          & \begin{tabular}[c]{@{}c@{}}\textbf{94.02}\\ \small±0.15\end{tabular} & \begin{tabular}[c]{@{}c@{}}93.53\\ \small±1.61\end{tabular}          & \begin{tabular}[c]{@{}c@{}}70.45\\ \small±0.24\end{tabular}          & \begin{tabular}[c]{@{}c@{}}\textbf{95.43}\\ \small±0.16\end{tabular} & \begin{tabular}[c]{@{}c@{}}68.34\\ \small±7.27\end{tabular}           & \begin{tabular}[c]{@{}c@{}}81.35\\ \small±0.63\end{tabular}          & \begin{tabular}[c]{@{}c@{}}82.60\\ \small±0.53\end{tabular}           & \begin{tabular}[c]{@{}c@{}}87.23\\ \small±0.09\end{tabular}          & 84.12           &                      &                               &  &  \\
TLM\scriptsize{(Medium)}       & 109M             & 1.21G        & 8.30E18          & \begin{tabular}[c]{@{}c@{}}\ul{93.96}\\ \small±0.18\end{tabular}          & \begin{tabular}[c]{@{}c@{}}\textbf{94.05}\\ \small±0.96\end{tabular} & \begin{tabular}[c]{@{}c@{}}70.90\\ \small±0.73\end{tabular}          & \begin{tabular}[c]{@{}c@{}}\ul{93.97}\\ \small±0.10\end{tabular}          & \begin{tabular}[c]{@{}c@{}}\ul{72.37}\\ \small±2.11\end{tabular}           & \begin{tabular}[c]{@{}c@{}}\ul{81.88}\\ \small±1.92\end{tabular}          & \begin{tabular}[c]{@{}c@{}}\ul{83.24}\\ \small±0.36\end{tabular}           & \begin{tabular}[c]{@{}c@{}}\ul{87.28}\\ \small±0.10\end{tabular}          & \ul{84.71}           &                      &                               &  &  \\

TLM\scriptsize{(Medium-20\%)}\tnote{3}  & 109M             & 0.18G        & 4.15E18          & \begin{tabular}[c]{@{}c@{}}93.78\\ \small±0.02\end{tabular}          & \begin{tabular}[c]{@{}c@{}}93.53\\ \small±0.00\end{tabular}           & \begin{tabular}[c]{@{}c@{}}\ul{71.11}\\ \small±0.05\end{tabular}          & \begin{tabular}[c]{@{}c@{}}93.20\\ \small±0.06\end{tabular}          & \begin{tabular}[c]{@{}c@{}}68.82\\ \small±3.56\end{tabular}         & \begin{tabular}[c]{@{}c@{}}80.35\\ \small±0.54\end{tabular}         & \begin{tabular}[c]{@{}c@{}}81.05\\ \small±0.07\end{tabular}          & \begin{tabular}[c]{@{}c@{}}87.00\\ \small±0.05\end{tabular}         & 83.58           &                      &                               &  &  \\
ISS\scriptsize{(Medium-scale)} & 109M             & 0.18G        & 4.15E18          & \begin{tabular}[c]{@{}c@{}}93.92\\ \small±0.08\end{tabular}          & \begin{tabular}[c]{@{}c@{}}93.53\\ \small±0.00\end{tabular}           & \begin{tabular}[c]{@{}c@{}}\textbf{71.51}\\ \small±0.31\end{tabular} & \begin{tabular}[c]{@{}c@{}}93.61\\ \small±0.06\end{tabular}          & \begin{tabular}[c]{@{}c@{}}\textbf{73.42}\\ \small±0.58\end{tabular} & \begin{tabular}[c]{@{}c@{}}\textbf{82.20}\\ \small±0.40\end{tabular} & \begin{tabular}[c]{@{}c@{}}\textbf{83.42}\\ \small±0.11\end{tabular} & \begin{tabular}[c]{@{}c@{}}\textbf{87.30}\\ \small±0.02\end{tabular} & \textbf{84.86}  &                      & \multicolumn{1}{c}{\textbf{}} &  &  \\ \cline{1-13}
RoBERTa-large    & 355M             & 160G          & 4.36E21          & \begin{tabular}[c]{@{}c@{}}\textbf{94.30}\\ \small±0.23\end{tabular} & \begin{tabular}[c]{@{}c@{}}\textbf{95.16}\\ \small±0.00\end{tabular} & \begin{tabular}[c]{@{}c@{}}70.73\\ \small±0.62\end{tabular}          & \begin{tabular}[c]{@{}c@{}}\textbf{96.20}\\ \small±0.19\end{tabular}  & \begin{tabular}[c]{@{}c@{}}72.80\\ \small±0.62\end{tabular}            & \begin{tabular}[c]{@{}c@{}}82.62\\ \small±0.68\end{tabular}          & \begin{tabular}[c]{@{}c@{}}\textbf{84.62}\\ \small±0.50\end{tabular}  & \begin{tabular}[c]{@{}c@{}}\textbf{87.53}\\ \small±0.13\end{tabular} & \textbf{85.50}   &                      &                               &  &  \\
TLM\scriptsize{(Large) }\tnote{4}       & 109M             & 3.64G        & 2.33E19          & \begin{tabular}[c]{@{}c@{}}94.15\\ \small±0.01\end{tabular}          & \begin{tabular}[c]{@{}c@{}}\ul{93.92}\\ \small±0.72\end{tabular}          & \begin{tabular}[c]{@{}c@{}}\ul{71.83}\\ \small±0.11\end{tabular}          & \begin{tabular}[c]{@{}c@{}}94.44\\ \small±0.10\end{tabular}          & \begin{tabular}[c]{@{}c@{}}\ul{74.18}\\ \small±0.29\end{tabular}           & \begin{tabular}[c]{@{}c@{}}\ul{82.77}\\ \small±0.72\end{tabular}          & \begin{tabular}[c]{@{}c@{}}\ul{83.60}\\ \small±0.08\end{tabular}          & \begin{tabular}[c]{@{}c@{}}\ul{87.49}\\ \small±0.02\end{tabular}          & 85.31           &                      &                               &  &  \\
TLM\scriptsize{(Large-20\%)}\tnote{3}   & 109M             & 0.72G       & 8.30E18          & \begin{tabular}[c]{@{}c@{}}93.79\\ \small±0.31\end{tabular}          & \begin{tabular}[c]{@{}c@{}}92.72\\ \small±0.783\end{tabular}         & \begin{tabular}[c]{@{}c@{}}71.50\\ \small±0.28\end{tabular}          & \begin{tabular}[c]{@{}c@{}}94.49\\ \small±0.04\end{tabular}          & \begin{tabular}[c]{@{}c@{}}73.42\\ \small±1.75\end{tabular}          & \begin{tabular}[c]{@{}c@{}}81.77\\ \small±0.54\end{tabular}          & \begin{tabular}[c]{@{}c@{}}82.63\\ \small±0.11\end{tabular}           & \begin{tabular}[c]{@{}c@{}}87.36\\ \small±0.10\end{tabular}          & 84.71           &                      &                               &  &  \\
ISS\scriptsize{(Large-scale)}  & 109M             & 0.72G       & 8.30E18          & \begin{tabular}[c]{@{}c@{}}\ul{94.22}\\ \small±0.04\end{tabular}          & \begin{tabular}[c]{@{}c@{}}93.53\\ \small±0.00\end{tabular}           & \begin{tabular}[c]{@{}c@{}}\textbf{72.27}\\ \small±0.20\end{tabular} & \begin{tabular}[c]{@{}c@{}}\ul{94.57}\\ \small±0.06\end{tabular}          &\begin{tabular}[c]{@{}c@{}}\textbf{74.53}\\ \small±1.38\end{tabular}  & \begin{tabular}[c]{@{}c@{}}\textbf{83.12}\\ \small±0.16\end{tabular} & \begin{tabular}[c]{@{}c@{}}83.31\\ \small±0.36\end{tabular}           & \begin{tabular}[c]{@{}c@{}}87.41\\ \small±0.02\end{tabular}          & \ul{85.36}     &                      &                               &  &  \\ \cline{1-13}
\end{tabular}
\end{threeparttable}
}
\begin{tablenotes}
    \scriptsize
    \item{1} For ISS, data size is reported by averaging over eight tasks. 
    \item{2} The training compute (FLOPs) is calculated by (6 × Training Tokens × Parameter Size) as in \citet{kaplan2020scaling}. 
    \item{3} ISS utilizes 20\% of the TLM size data, so we implemented the TLM model with the same size version.
    \item{4} For a fair comparison, we implement TLM\tiny{(Large)} with BERT base and TLM large scale dataset. 
  \end{tablenotes}
\caption{Evaluation results for ISS at three different training scales. For each task, we report the average F1 score across three random seeds with standard deviations as subscripts. We also show the number of parameters, the total training compute (FLOPs), and the size of training corpus for comparison.}
\label{table:mainResultsScratch}
\vspace{-3mm}
\end{table*}

\textbf{Datasets.}
Following the setting of \citet{gururangan-etal-2020-dont, yao2022nlp}, we conduct experiments on eight tasks covering four domains, including biomedical science, computer science, news, and reviews.
The tasks represent both high- and low-resource ($\leq$ 5K samples) settings, including CHEMPROT \cite{kringelum2016chemprot}, RCT \cite{dernoncourt-lee-2017-pubmed}, ACL-ARC \cite{jurgens-etal-2018-measuring}, SCIERC \cite{luan-etal-2018-multi}, HyPERPARTISAN \cite{kiesel-etal-2019-semeval}, AGNEws \cite{zhang2015character}, HELPFULNESS \cite{mcauley2015image}, IMDB \cite{maas-etal-2011-learning}.
Table \ref{table:datasets} reports the statistic results of various target datasets.
Similar to TLM \cite{yao2022nlp}, we collect two pretraining corpora that respectively match the original corpora of BERT and RoBERTa. We name them $\mathcal{C}_{BERT}$ and $\mathcal{C}_{RoBERTa}$, respectively.

\begin{figure*}[htbp]
\vspace{-12mm}
\centering
\subfigure[Helpfullness ]{
\begin{minipage}[t]{0.245\linewidth}
\centering
\includegraphics[width=1.7in]{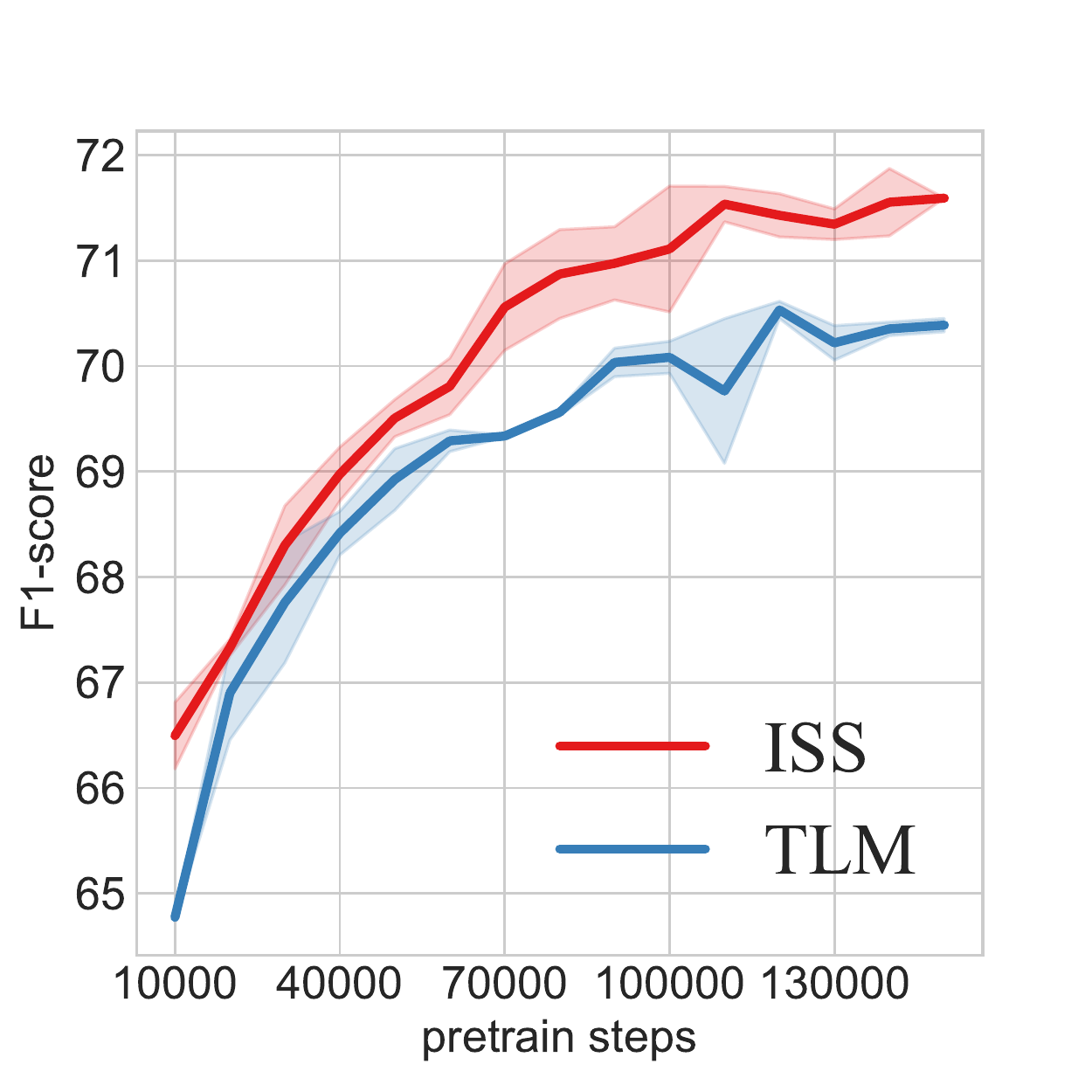}
\end{minipage}%
}%
\subfigure[Chemprot]{
\begin{minipage}[t]{0.245\linewidth}
\centering
\includegraphics[width=1.7in]{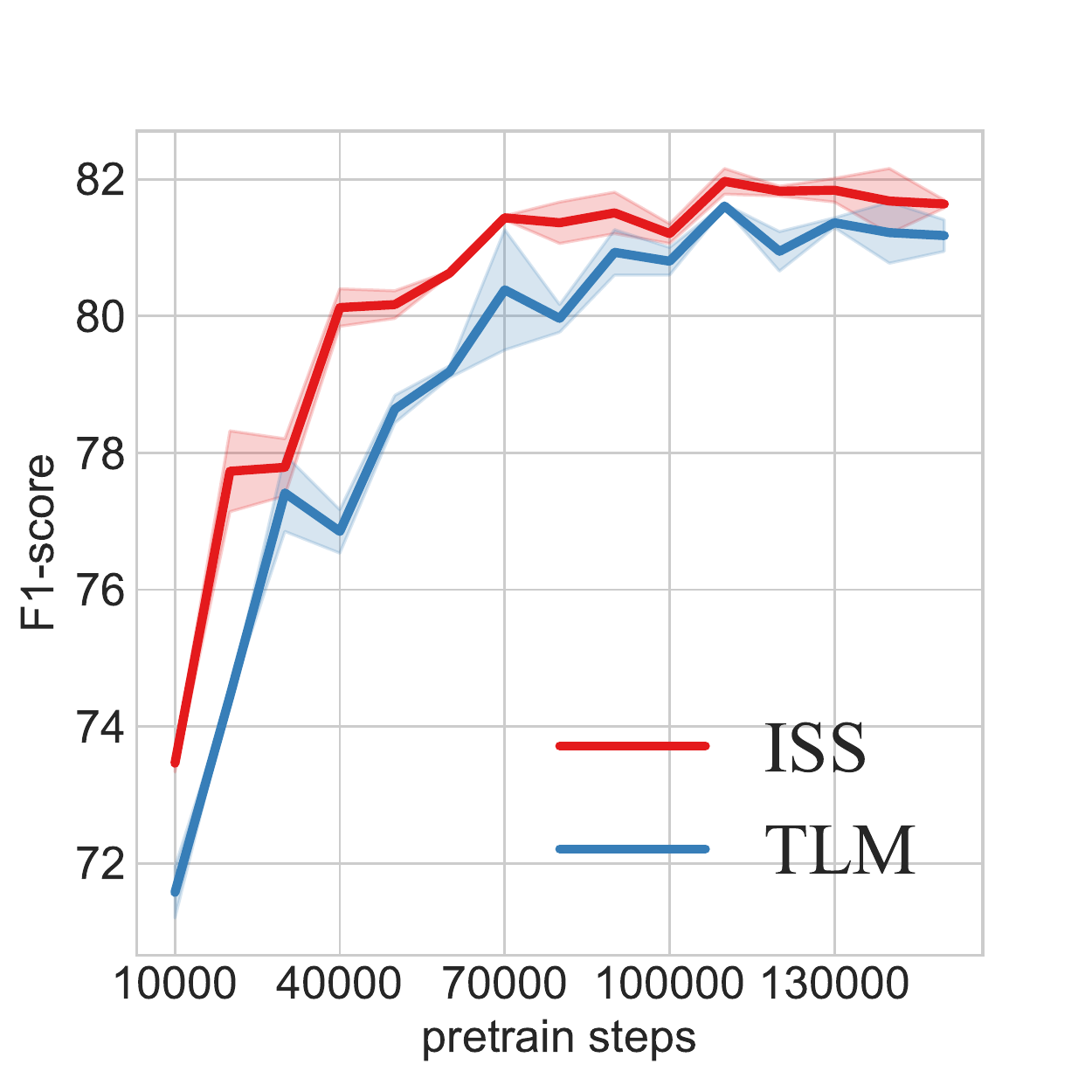}
\end{minipage}%
}%
\subfigure[SciERC]{
\begin{minipage}[t]{0.245\linewidth}
\centering
\includegraphics[width=1.7in]{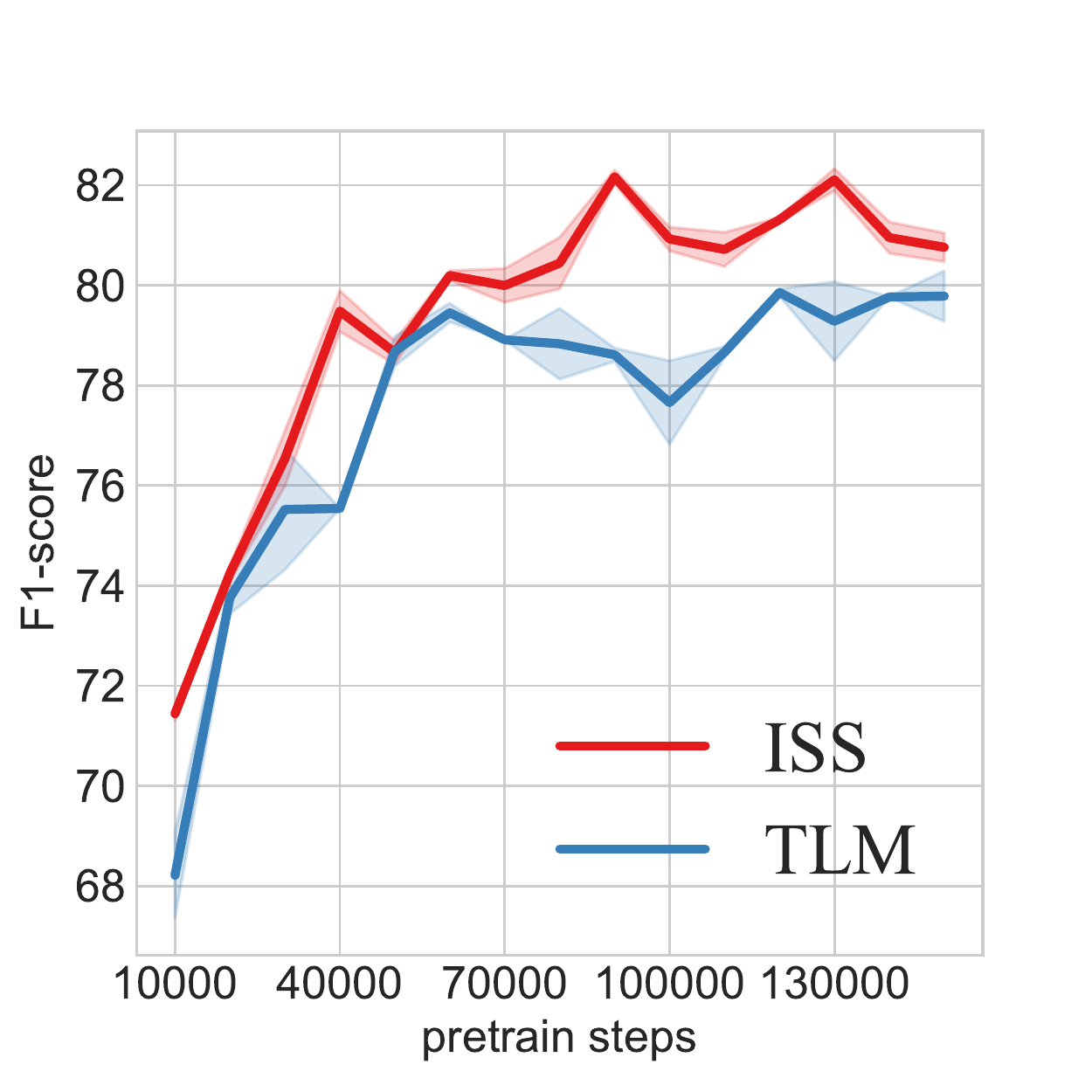}
\end{minipage}
}%
\subfigure[ACL-ARC]{
\begin{minipage}[t]{0.245\linewidth}
\centering
\includegraphics[width=1.7in]{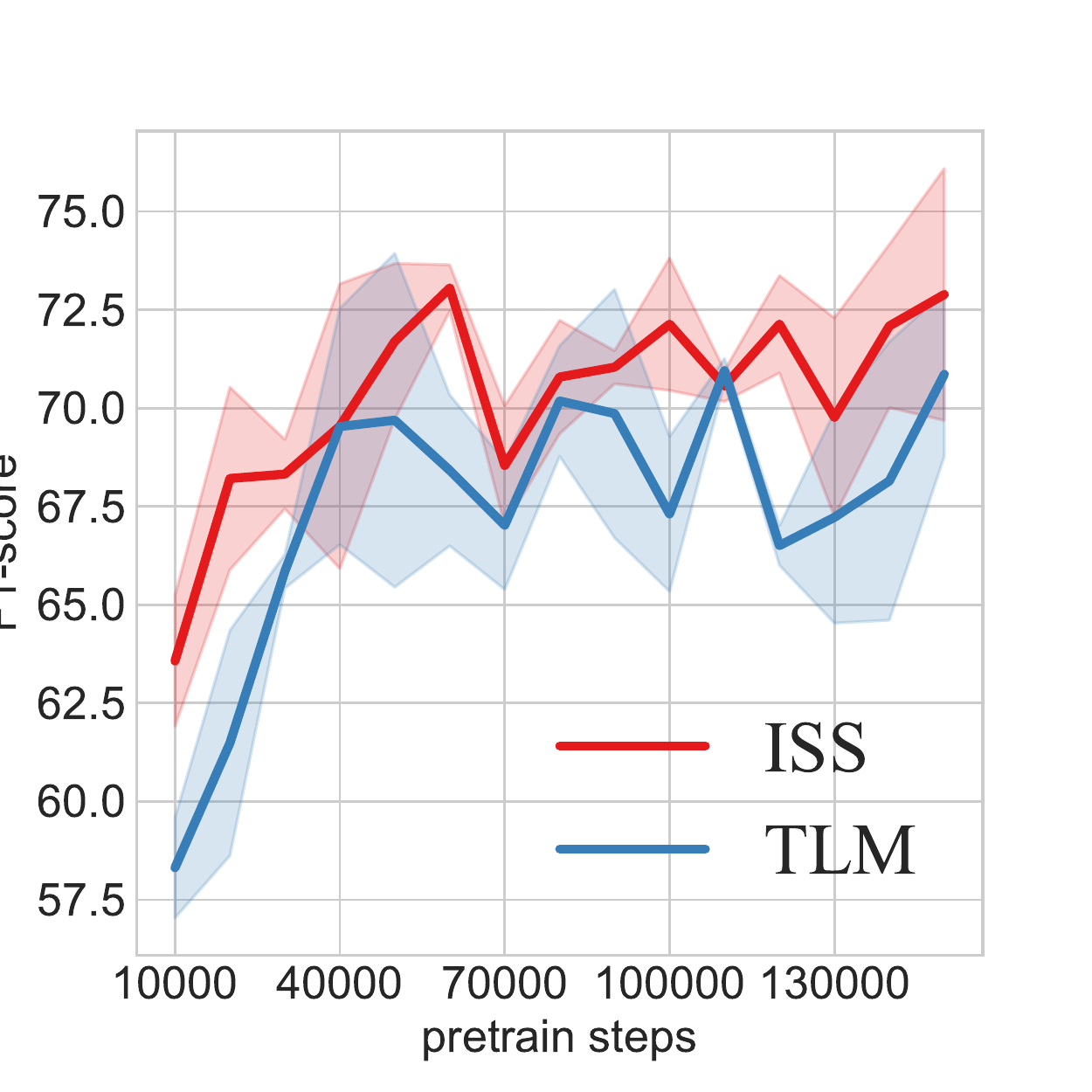}
\end{minipage}
}%

\vspace{-2mm}
  \caption{Comparison of ISS and TLM at different pretraining steps. 
  The experiments were conducted on the small-scale dataset, and notably, the data scale of TLM was five times larger than ours.}
 \label{fig:DifferentStep}
 \vspace{-2mm}
\end{figure*}

\textbf{Baselines.}
We focus on comparison with general PLMs and TLM. 
Following \citet{yao2022nlp}, we finetuned both BERT \cite{devlin-etal-2019-bert} and RoBERTa \cite{liu2019roberta} of base and large scales as our baselines.
And we finetuned the released TLM models as baselines.

\textbf{Evaluation Strategy.}
The results of the experiment are the average performance of three random seeds with the standard deviations.
Following \citet{gururangan-etal-2020-dont}, we report the test micro-F1 for ChemProt and RCT, and macro-F1 for the rest of datasets. 
Following TLM \cite{yao2022nlp}, we set three pretraining scales, namely $small$, $medium$, and $large$ scales. 
Differently, at the same scale, our method only utilizes 20\% size of the TLM data. More detailed settings are shown in Table \ref{DetailedSettingsScratch} in Appendix.

\textbf{Training Details.}
We utilize the randomly initialized BERT of base scale as our starter models. 
We mostly follow optimization, and hyper-parameters choices used in \citet{yao2022nlp}.
All experiments were conducted on 4 NVIDIA GeForce RTX 3090 GPUs.
Detailed hyper-parameters are provided in Table \ref{DetailedSettingsScratch} in Appendix.

\subsection{Further Pretraining}

\textbf{Datasets.}
We perform further pretraining in biomedical science and computer science domains. 
Specifically, we conduct experiments on four datasets, including CHEMPROT \cite{kringelum2016chemprot}, RCT \cite{dernoncourt-lee-2017-pubmed}, ACL-ARC \cite{jurgens-etal-2018-measuring}, SCIERC \cite{luan-etal-2018-multi}. 
For the pretraining stage, we collect the unlabeled datasets from S2ORC \cite{lo-etal-2020-s2orc}.

\textbf{Baselines.}
We select general PTMs \cite{devlin-etal-2019-bert,liu2019roberta} and domain-specific further pretraining models \cite{lee2020biobert, beltagy-etal-2019-scibert, gururangan-etal-2020-dont} as our baselines. 
Finetuning on the end-task occurs after further pretraining on domain unlabeled corpora.




\textbf{Evaluation Strategy.}
Similar to pretraining from scratch, we report the average performance across three random seeds.
And we report the micro-F1 for ChemProt and RCT, and macro-F1 for ACL-ARC and SCIERC. 

\textbf{Training Details.}
In this setting, we perform further pretraining on off-the-shelf pretrained models, such as BERT and RoBERTa.
All experiments were conducted on 4 NVIDIA GeForce RTX 3090 GPUs.
Detailed hyper-parameters are provided in Table \ref{table:furtherSettings} in Appendix.

\section{Experimental Results}
In this section, we will discuss the results of comparing our methods against other baselines. 

\subsection{Pretraining from Scratch}
Table \ref{table:mainResultsScratch} shows the main results of ISS with the according TLM and PLMs baselines at three different scales.
The followings are the related comparison and analysis we conducted:
\textbf{1)} ISS could achieve results that are better than or comparable to the PLM baselines with significant reductions in FLOPs and the size of training data. At the large scale, ISS achieves comparable results to RoBERTa-large, with an average of 0.19\% of FLOPs and 0.45\% of the training corpus. At the small and medium scales, ISS improves the performance by 0.29 and 0.74 points on average respectively; 
\textbf{2)} At the same data scale, ISS significantly outperforms TLM, which indicates that task label information is crucial. And the influence-based subset selection can select more influential pertaining samples;
\textbf{3)} ISS could offer limited performance gains on high-resource datasets. It demonstrates that the influence of the pretraining samples would be decreased as the task data grows sufficiently.

\subsection{Further Pretraining}
\begin{table}[t]
\vspace{-2mm}
\renewcommand\arraystretch{1.8}
\setlength{\tabcolsep}{1mm}
 \resizebox{1\columnwidth}{!}{
\begin{threeparttable}
\begin{tabular}{l|ccc|cc|cc|c}
\hline
\multirow{2}{*}{\textbf{Model}} & \multirow{2}{*}{\textbf{Param}} & \multirow{2}{*}{\textbf{Data}} & \multirow{2}{*}{\textbf{FLOPs}} & \multicolumn{2}{c|}{\textbf{BIOMED}}                                                                                            & \multicolumn{2}{c|}{\textbf{CS}}                                                                                                        & \multirow{2}{*}{\textbf{Avg.}} \\
                       &                                 &                                &                                 & RCT                                                   & Chem                                                           & ACL                                                           & SciERC                                                         &                                \\ \hline
BERT-Base              & 109M                            & 16G                            & 2.79E19                         & 87.00                                                 & 81.94                                                          & 69.45                                                         & 80.98                                                          & 79.84                          \\
RoBERTa-base           & 125M                            & 160G                           & 1.54E21                         & 87.23                                                 & 82.60                                                          & 68.34                                                         & 81.35                                                          & 79.88                          \\ \hline
SciBERT                & 109M                            & 15G                          & 2.65E19                         & -                                                     & 83.64                                                          & 70.98                                                         & 79.97                                                          & -                              \\

BioBERT                & 109M                            & 96G                            & 1.80E20                         & -                                                     & 76.46                                                          & -                                                             & -                                                              & -                              \\

DAPT          & 125M                            & 47G                            & 1.58E18                         & 87.6                                         & 84.2                                                           & 75.4                                                          & 80.8                                                           & 82.00                          \\
DAPT+TAPT                & 125M                            & 47G                            & 1.77E18                         & \textbf{87.8}                                                     & 84.4                                                          & 75.6                                                             & 81.3                                                              & 82.28                              \\
\hline
ISS-DAPT\scriptsize{(BERT)}        & 109M                            & 1.7G                             & 6.9E17                         & \begin{tabular}[c]{@{}c@{}}87.36\\ \small$\pm$0.02\end{tabular} & \begin{tabular}[c]{@{}c@{}}83.90\\ \small$\pm$0.10\end{tabular}          & \begin{tabular}[c]{@{}c@{}}76.06\\ \small $\pm$0.70\end{tabular}         & \begin{tabular}[c]{@{}c@{}}\textbf{83.91}\\ \small$\pm$0.38\end{tabular} & 82.81                          \\
ISS-DAPT\scriptsize{(RoBERTa)}      & 125M                            & 1.7G                             & 7.9E17                         & \begin{tabular}[c]{@{}c@{}}87.57\\ \small$\pm$0.06\end{tabular} & \begin{tabular}[c]{@{}c@{}}\textbf{84.88}\\ \small$\pm$0.10\end{tabular} & \begin{tabular}[c]{@{}c@{}}\textbf{76.70}\\ \small$\pm$0.25\end{tabular} & \begin{tabular}[c]{@{}c@{}}82.23\\ \small$\pm$0.30\end{tabular}          & \textbf{82.85}                 \\ \hline
\end{tabular}
\end{threeparttable}
}
\caption{Evaluation results for ISS in further pretraining. We report the average F1 score across three random seeds with standard deviations as subscripts.}
\label{table:furtherResults}
\vspace{-2mm}

\end{table}

We compared ISS with other domain-specific further pretraining methods. 
Differently, we initialize the network with off-the-shelf pretrained models to provide initialization and select influential subsets from the domain corpus.
Table \ref{table:furtherResults} shows the main results.
In conclusion, our method outperforms all the baselines,
with significant reductions in FLOPs and the size of training data by one order of magnitude or more. It proves our approach is feasible.


\subsection{Comparison of Pretraining Steps}
To validate the effect of pretraining steps, we compare the performance of ISS with TLM at different pretraining steps.
The test results on the four tasks with different 
 pretraining steps are shown in Figure  \ref{fig:DifferentLayers}.
We observe that ISS could achieve the best performance with fewer steps on most of the datasets.

\begin{table}[t]
\vspace{-2mm}
\centering
\small
\renewcommand\arraystretch{1.4}
\setlength{\tabcolsep}{1.5mm}
 \resizebox{1\columnwidth}{!}{

 \begin{tabular}{l|cccccc}
\hline
\multirow{2}{*}{} & \multicolumn{2}{c}{AGNews}                                                                                                      & \multicolumn{2}{c}{SciERC}                                                                                                      & \multicolumn{2}{c}{Chemprot}                                                                                                    \\
                  & ISS                                                            & TLM                                                            & ISS                                                            & TLM                                                            & ISS                                                            & TLM                                                            \\ \hline
10\%              & \begin{tabular}[c]{@{}c@{}}94.34\\ \small±0.08\end{tabular}          & \begin{tabular}[c]{@{}c@{}}94.08\\ \small±0.07\end{tabular}          & \begin{tabular}[c]{@{}c@{}}80.82\\ \small±0.41\end{tabular}          & \begin{tabular}[c]{@{}c@{}}81.41\\ \small±0.16\end{tabular}          & \begin{tabular}[c]{@{}c@{}}80.80\\ \small±0.34\end{tabular}          & \begin{tabular}[c]{@{}c@{}}80.15\\ \small±0.32\end{tabular}          \\
20\%              & \begin{tabular}[c]{@{}c@{}}\textbf{94.40}\\ \small±0.06\end{tabular} & \begin{tabular}[c]{@{}c@{}}94.16\\ \small±0.09\end{tabular} & \begin{tabular}[c]{@{}c@{}}\textbf{83.70}\\ \small±0.31\end{tabular} & \begin{tabular}[c]{@{}c@{}}81.21\\ \small±0.44\end{tabular}          & \begin{tabular}[c]{@{}c@{}}\textbf{82.82}\\ \small±0.41\end{tabular} & \begin{tabular}[c]{@{}c@{}}81.51\\ \small±0.55\end{tabular}          \\
40\%              & \begin{tabular}[c]{@{}c@{}}94.14\\ \small±0.05\end{tabular}          & \begin{tabular}[c]{@{}c@{}}94.05\\ \small±0.18\end{tabular}          & \begin{tabular}[c]{@{}c@{}}83.16\\ \small±0.07\end{tabular}          & \begin{tabular}[c]{@{}c@{}}82.48\\ \small±0.43\end{tabular}          & \begin{tabular}[c]{@{}c@{}}81.98\\ \small±0.14\end{tabular}          & \begin{tabular}[c]{@{}c@{}}81.75\\ \small±0.04\end{tabular}          \\
60\%              & \begin{tabular}[c]{@{}c@{}}94.08\\ \small±0.02\end{tabular}          & \begin{tabular}[c]{@{}c@{}}94.07\\ \small±0.09\end{tabular}          & \begin{tabular}[c]{@{}c@{}}82.51\\ \small±0.29\end{tabular}          & \begin{tabular}[c]{@{}c@{}}\textbf{83.05}\\ \small±0.20\end{tabular} & \begin{tabular}[c]{@{}c@{}}82.08\\ \small±0.22\end{tabular}          & \begin{tabular}[c]{@{}c@{}}81.80\\ \small±0.41\end{tabular}          \\
80\%              & \begin{tabular}[c]{@{}c@{}}94.17\\ \small±0.04\end{tabular}          & \begin{tabular}[c]{@{}c@{}}\textbf{94.27}\\ \small±0.09\end{tabular}          & \begin{tabular}[c]{@{}c@{}}81.71\\ \small±0.24\end{tabular}          & \begin{tabular}[c]{@{}c@{}}81.75\\ \small±0.15\end{tabular}          & \begin{tabular}[c]{@{}c@{}}81.83\\ \small±0.30\end{tabular}          & \begin{tabular}[c]{@{}c@{}}\textbf{81.86}\\ \small±0.47\end{tabular} \\ \hline

\end{tabular}

}
\caption{Results on the development set with different data scales. 
}
\vspace{-2mm}

\label{table:different-size}
\end{table}

\subsection{Subset Size for Pretraining}
To compare the performance at different data scales, 
we extracted subsets from the TLM small-scale corpus at different scales via ISS and TLM, respectively.
The results are shown in Table \ref{table:different-size}. 
We can observe that the performance of TLM becomes better as the dataset grows, but the best results are still lower than those of our method. 
In ISS, the F1-score would reach the top at the 20\%-40\% scale and gradually decrease as the data size grows. 
We believe that as the dataset expands, task-irrelevant or noisy data is added.

\subsection{Last Better than First}
\begin{figure}[t]
\centering
\subfigure[Chemprot]{
\begin{minipage}[t]{0.47\linewidth}
\centering
\includegraphics[width=1.4in]{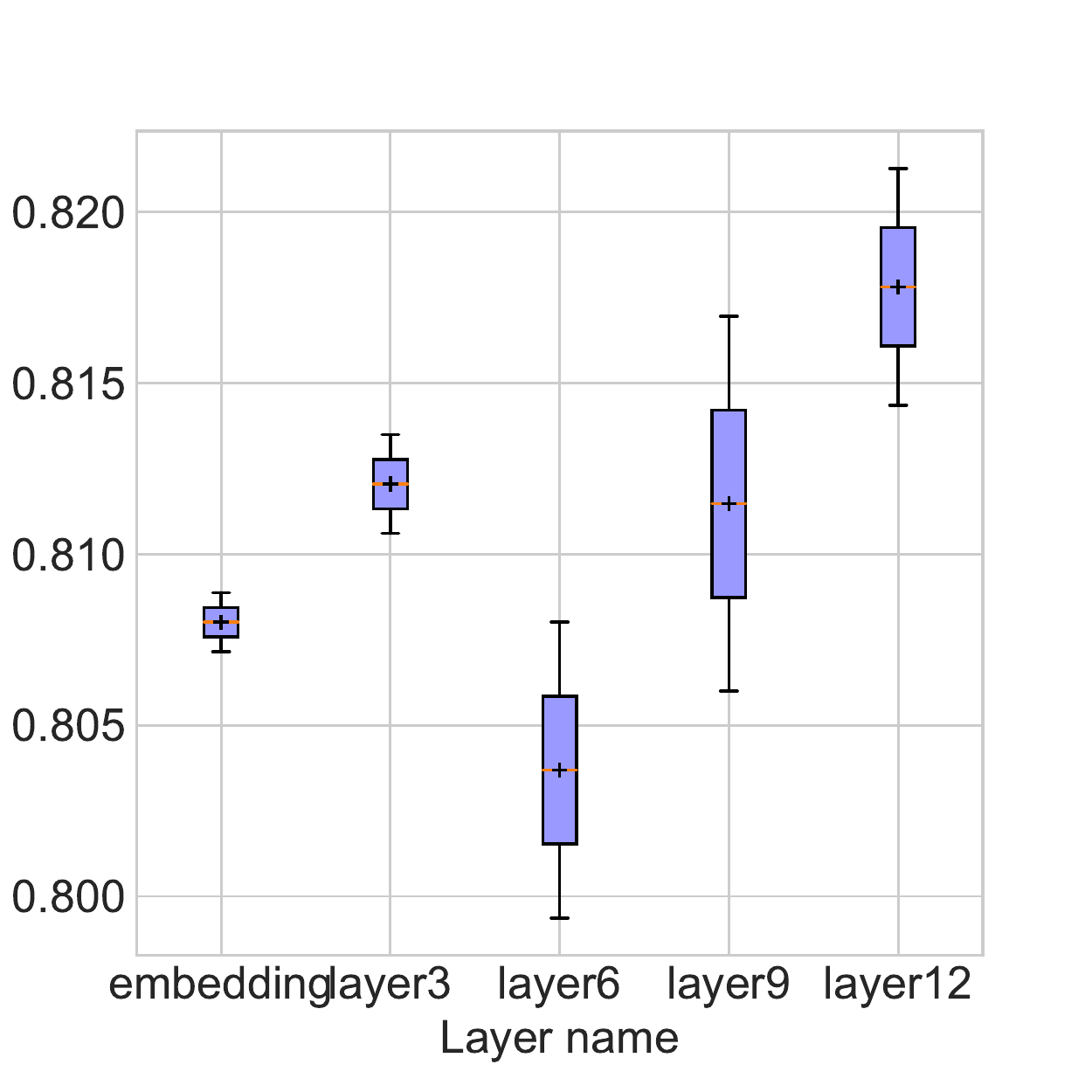}
\label{fig:layer-chemprot}
\end{minipage}%
}%
\subfigure[SciERC]{
\begin{minipage}[t]{0.47\linewidth}
\centering
\includegraphics[width=1.4in]{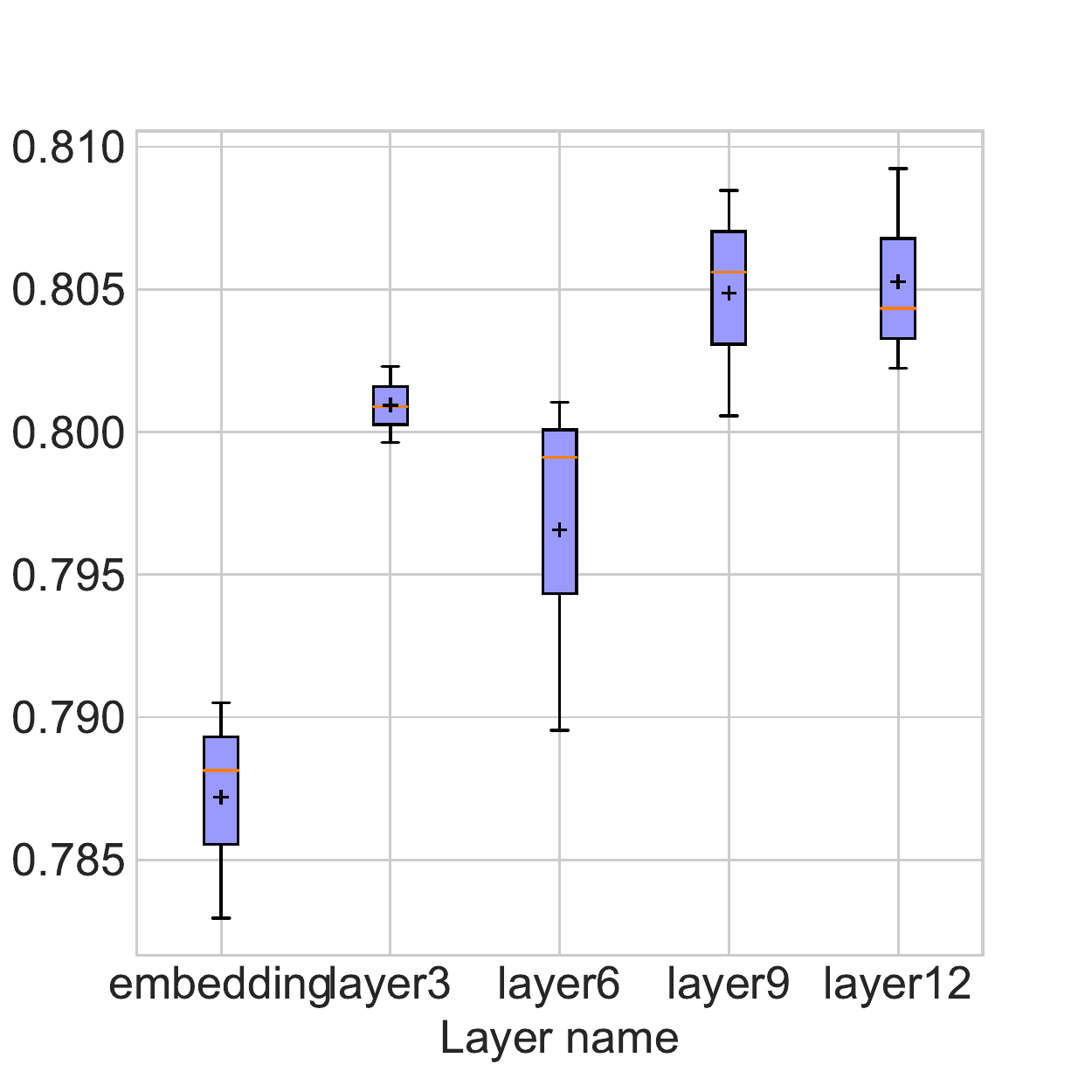}
\label{fig:layer-sciie}
\end{minipage}%
}%
\vspace{-2mm}
  \caption{F1-score results of ISS with gradients of different layers (i.e., Embedding layer, 3/6/9/12-th Transformer Block) over Chemprot and SciERC. }
 \label{fig:DifferentLayers}
 \vspace{-3mm}
\end{figure}

\begin{table}[t]
\vspace{0mm}
\centering
\renewcommand\arraystretch{1.4}
\setlength{\tabcolsep}{1mm}
 \resizebox{1\columnwidth}{!}{
\begin{tabular}{l|ccccl}
\cline{1-5}
\multirow{2}{*}{} & \multirow{2}{*}{Related-Label} & \multirow{2}{*}{PMI} & \multicolumn{2}{c}{AGNews} & \multicolumn{1}{c}{} \\
                  &                                &                      & ISS\small{(small)} /\%    & TLM\small{(small)} /\%   & \multicolumn{1}{c}{} \\ \cline{1-5}
immigration       & World                          & 1.341                & 0.0072       & 0.0070      &                      \\
policy            & World                          & 1.187                & 0.0493       & 0.0401      &                      \\
china             & World                          & 0.382                & 0.0836       & 0.0695      &                      \\
medals            & Sports                         & 1.400                & 0.0139       & 0.0136      &                      \\
golds             & Sports                         & 1.400                & 0.0009       & 0.0008      &                      \\
sports            & Sports                         & 1.293                & 0.0459       & 0.0454      &                      \\
financial         & Business                       & 1.054                & 0.0717       & 0.0567      &                      \\
commerce          & Business                       & 0.844                & 0.0097       & 0.0081      &                      \\
business          & Business                       & 0.710                & 0.1170       & 0.0952      &                      \\
automation        & Sci/Tech                       & 1.420                & 0.0043       & 0.0028      &                      \\
internet          & Sci/Tech                       & 1.224                & 0.0729       & 0.0524      &                      \\
technology        & Sci/Tech                       & 1.115                & 0.0864       & 0.0661      &                      \\ \cline{1-5}
\end{tabular}
 
 }
 \caption{Comparison of the frequency of task influential words in different subsets.}
 \label{table:PMI-freq}
 \vspace{-3mm}
 \end{table}
 
As explained in Section \ref{implementDetails}, the last layer of gradients of the model encoder is only considered to speed up the computation.
We have studied the relationship between the gradients at the different layers used in ISS and the corresponding performances.
Table \ref{fig:DifferentLayers} shows the results on Chemprot and SciERC.
We can observe that the closer the layer, to the task head, the better the selected subset works.
The phenomena suggest that different layers in the language model can capture different information, with layers closer to the task head learning more information about the task.

Table \ref{table:ISS-time} shows the times required by ISS calculating influences at the different layers.
Overall, the time cost of selecting a subset is negligible compared to pretraining.
In addition, the computational speed based on the last layer would be nearly double, compared to that at the embedding layer.

\begin{table}[t!]
\small
\centering
\renewcommand\arraystretch{1.5}
 \resizebox{1\columnwidth}{!}{
\begin{tabular}{lcc}
\hline
\multicolumn{1}{l|}{\multirow{2}{*}{{Layer name}}} & \multicolumn{2}{c}{{Cost times}}     \\
\multicolumn{1}{l|}{}                                     & {Small}       & {Large}       \\ \hline
\multicolumn{1}{l|}{Embedding}                            & 2.0 hours               & 5.2 hours               \\
\multicolumn{1}{l|}{3-th Transformer}                      & 1.8 hours            & 4.8 hours            \\
\multicolumn{1}{l|}{6-th Transformer}                      & 1.6 hours            & 4.4 hours            \\
\multicolumn{1}{l|}{9-th Transformer}                      & 1.4 hours            & 4.0 hours               \\
\multicolumn{1}{l|}{12-th Transformer}                     & 1.1 hours            & 3.6 hours            \\ \hline
\end{tabular}
 
 }
 \caption{Comparison of the speed of computing influences using different layers. The experiments were conducted on Chemport dataset.}
  \label{table:ISS-time}
 \vspace{-3mm}
\end{table}

\section{Analysis}
\subsection{Visualization of Pretrained Model}

\begin{figure*}[htbp]
\vspace{-12mm}
\centering
\subfigure[ISS]{
\begin{minipage}[t]{0.325\linewidth}
\centering
\includegraphics[width=2.1in]{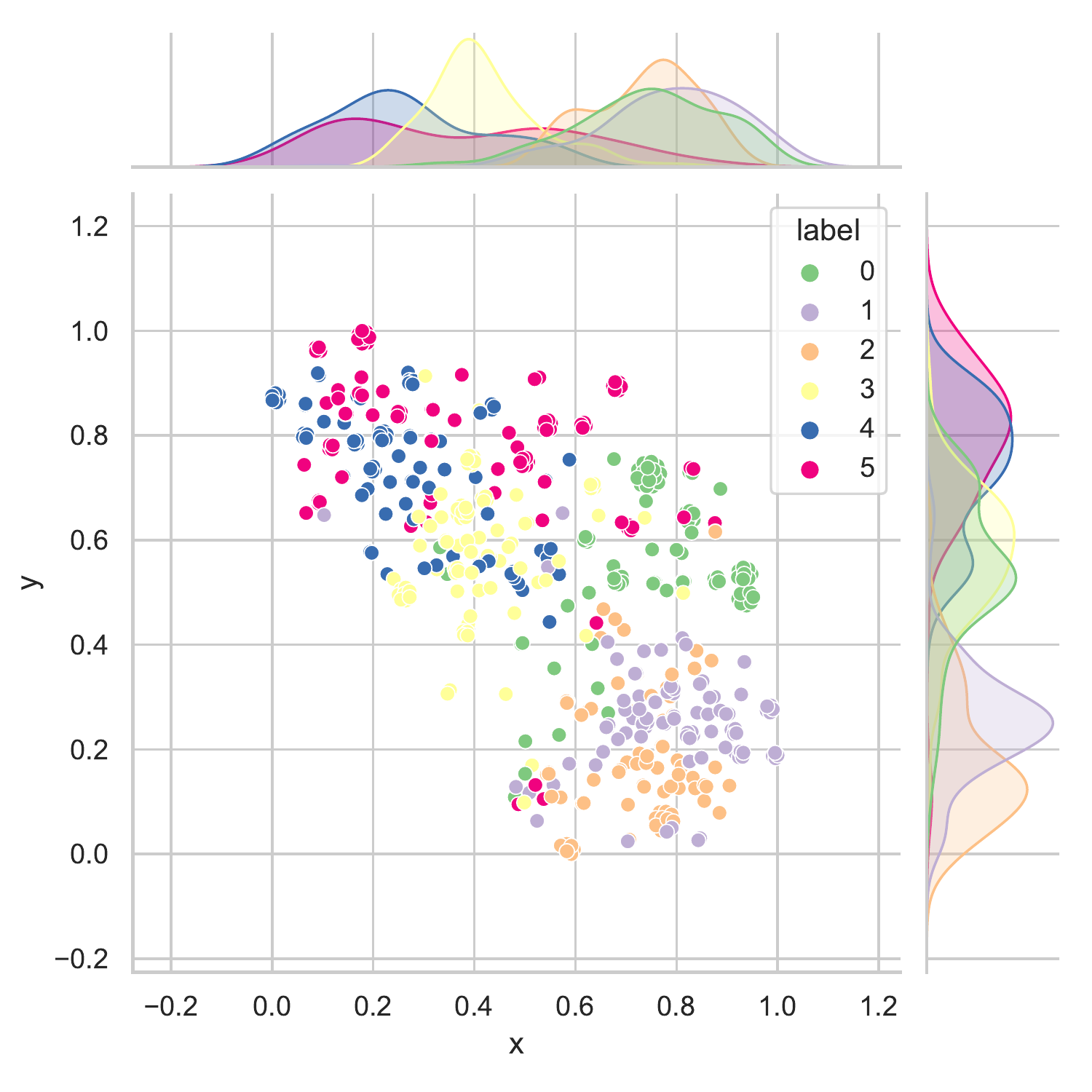}
\label{isp-rep}
\end{minipage}%
\vspace{-2mm}
}%
\subfigure[BERT]{
\begin{minipage}[t]{0.325\linewidth}
\centering
\includegraphics[width=2.1in]{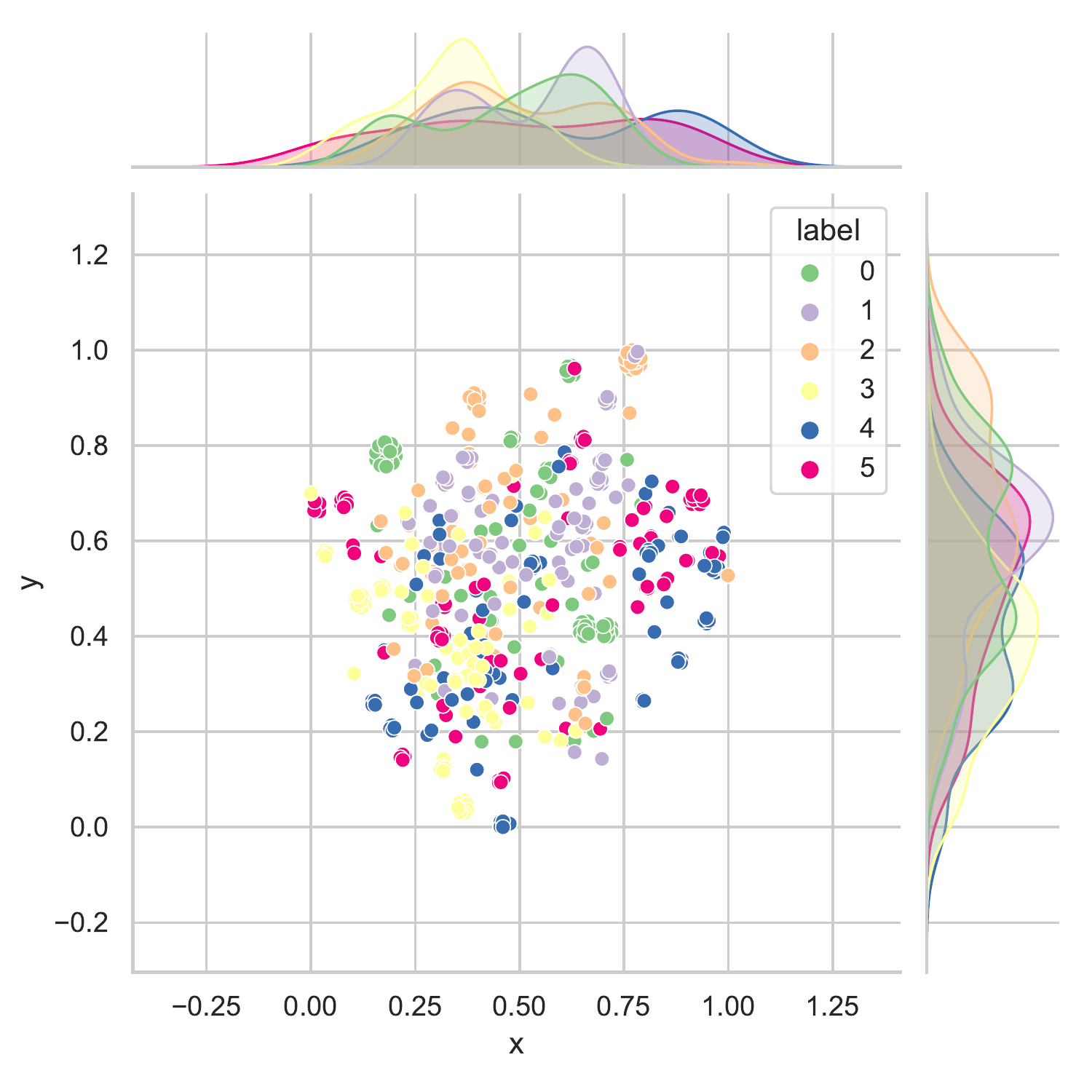}
\label{bert-rep}
\end{minipage}%
\vspace{-2mm}
}%
\subfigure[RoBERTa]{
\begin{minipage}[t]{0.325\linewidth}
\centering
\includegraphics[width=2.1in]{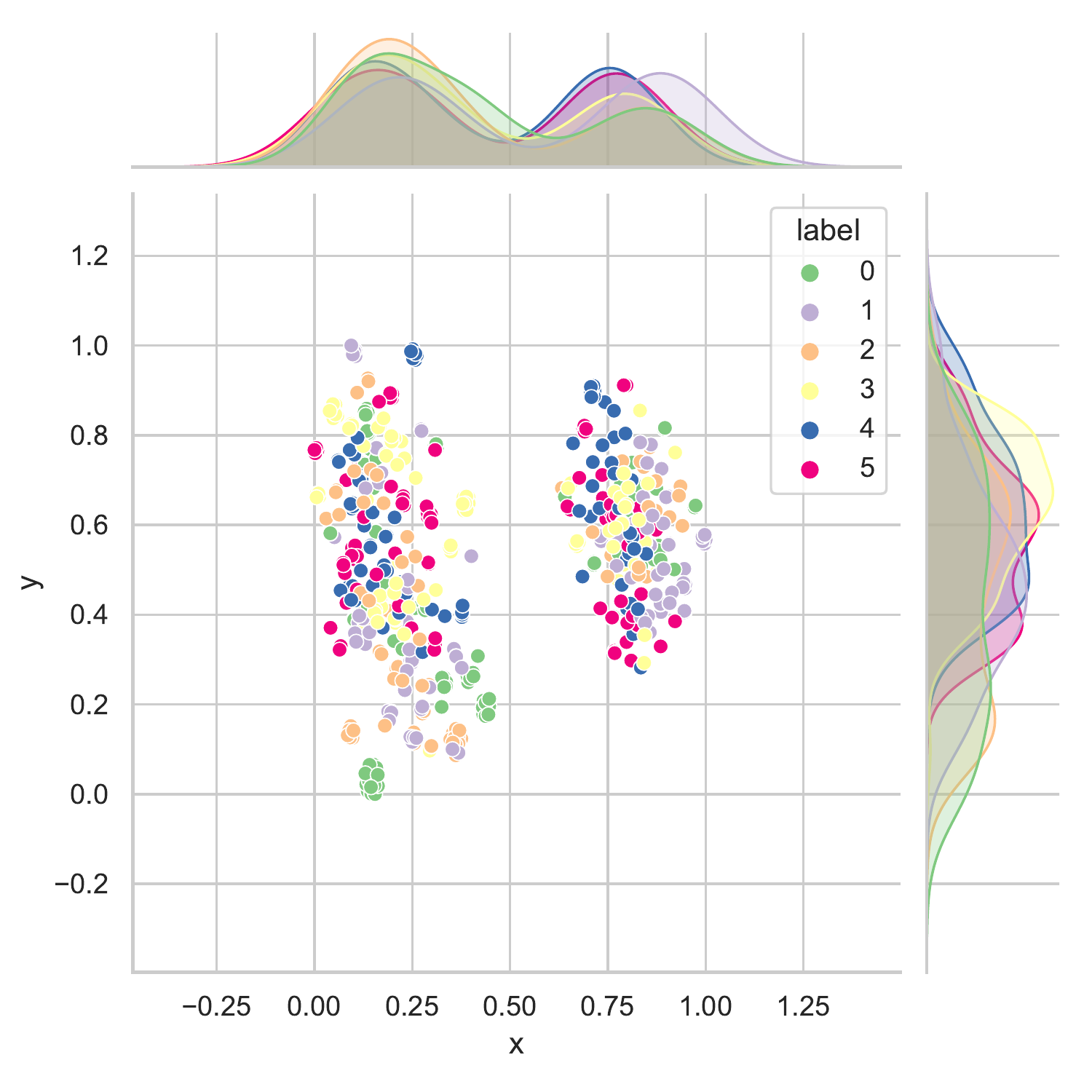}
\label{roberta-rep}
\end{minipage}
\vspace{-2mm}
}%
\centering
\vspace{-3mm}
\caption{ Visualization of sentence representation on Chemprot using the t-SNE algorithm \cite{van2008visualizing}. Each color denotes a class. 
}
\label{fig:Visualization}
\vspace{-3mm}
\end{figure*}

We visualize the task data on ISS-small, BERT, and RoBERTa, using the t-SNE algorithm \cite{van2008visualizing}. 
The results are shown in Figure \ref{fig:Visualization}.
We can observe that the different classes of deep features in ISS-small formed tighter clusters, suggesting that ISS provides better initialization for downstream tasks. 
In contrast, the features learned by BERT and Roberta are distributed respectively in separate clusters with overlapping parts that could not be distinguished.

\subsection{Analyzing of Task-influential Words}
We compute the point-wise mutual information (PMI) \cite{levy2014neural} between words and their corresponding labels in the task dataset. Briefly, PMI is used to measure the likelihood of two events occurring together, so the higher the PMI a word has, the more likely it is to be task-influential. We select words with high PMI as task-influential words, and compare their frequency in ISS-small and TLM-small datasets, respectively. As shown in Table \ref{table:PMI-freq}, the word frequency in the ISS-small dataset is higher than that in the TLM-small dataset. Thus, ISS may focus more on task-influential words.

\section{Related Work}
\subsection{Efficient Pretraining for PLMs}
Many attempts have been made to improve the efficiency of pretraining. 
Parallel architectures \cite{shazeer2018mesh, wang2020minilm} are commonly used in pretraining.
However, parallelism would not actually reduce computational costs in terms of FLOPs.
For most Transformer-based PTMs, as their input sequence goes longer, their efficiency is limited by the computation of attention weights.
\citet{choromanski2020rethinking} and \citet{wang2020linformer} design low-rank kernels to theoretically approximate the original attention weights.
\citet{child2019generating} and \citet{roy2021efficient} introduce sparsity into attention mechanisms by limiting the view of each token to a fixed size and separating tokens into several chunks.
ELECTRA \cite{clark2019electra} applies the replaced token detection which is more challenging. 
PMI-Masking \cite{levine2020pmi} selectively masks tokens based on their importance.
However, their improvements are limited, with less than an order of magnitude reduction in computational expenses (measured in FLOPs).
Orthogonal to these works, ISS investigates reducing training data redundancy by the influence of pretraining data points.

\subsection{Further Pretraning in NLP}
Continually pretraining can effectively improve PTMs’ performance on new domains or downstream tasks \cite{gururangan-etal-2020-dont}. 
To achieve it, most previous works continually optimize the pretrained model parameters on a large number of corpora collected from the target domain (e.g., scientific \cite{beltagy-etal-2019-scibert}, finance\cite{araci2019finbert} and bio-media \cite{lee2020biobert}).
However, it is computationally expensive to further pretrain the model on a large amount of unlabeled data and it may not be feasible to collect such a large scale of unlabeled data on certain domains.
In contrast, ISS does not need any additional domain data and only utilizes the general corpus.
In addition, our approach can also be employed for further pretraining, as we demonstrate in our experiments.

\subsection{Dataset Pruning}
Dataset pruning is closely related to the coreset selection methods \cite{mirzasoleiman2020coresets,agarwal2004approximating}, which try to identify the most representative training samples.
Several works \cite{ killamsetty2021grad, rebuffi2017icarl, toneva2018empirical} have studied dataset pruning for efficient training of deep learning models in supervised learning and active learning scenarios. 
Dataset pruning methods typically rely on a pre-defined criterion to compute a scalar score for each training example, e.g. the compactness \cite{rebuffi2017icarl}, diversity \cite{sener2017active}, and forgetfulness \cite{toneva2018empirical}, and then rank and select the training data according to the computed score.
Recently, \citet{yao2022nlp} proposed TLM for transfer learning, which retrieves a subset from the pretraining corpus that is more similar to the task corpus.
However, these methods are heuristic and lack of generalization guarantee, they also discard the influence interaction between the collected samples. 
Our proposed method overcomes these shortcomings.


\section{Conclusion}
In this paper, we propose Influential Subset Selection for language model, which aims to reduce the computational costs of pretraining from data level.
Specifically, we introduce influence function to measure the importance of each pretraining sample.
Moreover, we design a simple, efficient, gradient matching-based method for influence estimation, which significantly speeds up the estimation time.
Experiments on various datasets demonstrate that our method achieves comparable performance with PTMs, with a reduction of training FLOPs by three orders of magnitude.

\section*{Limitations}
There are two potential risks with our method.
First, ISS trades generality for efficiency by learning only task-specific representations. 
Consequently, it may not be suitable for other tasks.
Secondly, our method is hardly practical for few-shot or zero-shot learning, as few or no task data are available as anchor points.
These potential risks are left to future work.

\section*{Ethics Statement}
Pretraining from scratch and further pretraining such as DAPT need large-scale unlabeled corpus to learn general knowledge, which results in corresponding greenhouse emissions due to energy consumption \cite{strubell-etal-2019-energy}. However, as shown in Section 5, our new efficient algorithms greatly increase the data efficiency of PTMs, reducing these harms as well as the various harms associated with labor for data collection. Our work introduces a new subset selection algorithm but leverages pre-existing datasets and models. Overall, this work inherits some of the risks of the original work upon which it is implemented, (such as bias \cite{bender2021dangers} or privacy leakage \cite{carlini2021extracting}.


\bibliography{anthology,custom}
\bibliographystyle{acl_natbib}

\appendix
\renewcommand\thetable{\Alph{section}.\arabic{table}}    
\section{Detailed Experiment Settings}
\setcounter{table}{0}

Table \ref{DetailedSettingsScratch} lists the detailed hyperparameters of ISS at different scales for each task on the pre-training task. On each task, we perform a grid search for $B_p$ $\in$ \{1, 2, 4, 8\} and Batch size$\small{(task)}$ $\in$ \{1,2,4,8,16\} and adjust the training step, batch size, and sequence length to minimize the training cost while maintaining competitive performance.

Table \ref{table:furtherSettings} lists the detailed hyperparameters of ISS for each task on further pretraining task.


\begin{table*}[t!]
\large
\centering
\renewcommand\arraystretch{1.5}
\begin{threeparttable}
 \resizebox{2\columnwidth}{!}{
\begin{tabular}{llccccccccl}
\cline{1-10}
\multicolumn{1}{l|}{}                                                                        & \multicolumn{1}{l|}{\textbf{Hyper-Parameters}} & \textbf{AGNews}      & \textbf{Hyp.}        & \textbf{Help.}       & \textbf{IMDB}        & \textbf{ACL.}        & \textbf{SciERC}      & \textbf{Chem.}       & \textbf{RCT}         &                      \\ \cline{1-10}
\multicolumn{1}{l|}{\multirow{9}{*}{\begin{tabular}[c]{@{}l@{}}Small\\ Scale\end{tabular}}}  & \multicolumn{1}{l|}{$B_p$}      & 4                    & 1                    & 4                    & 4                    & 4                    & 4                    & 4                    & 4                    &                      \\
\multicolumn{1}{l|}{}                                                                        & \multicolumn{1}{l|}{$B_t$}          & 16                   & 1                    & 16                   & 16                   & 2                    & 4                    & 8                    & 16                   &                      \\
\multicolumn{1}{l|}{}                                                                        & \multicolumn{1}{l|}{Source Corpus\tnote{1}}             & $\mathcal{C}_{TLM-small}$     & $\mathcal{C}_{TLM-small}$     & $\mathcal{C}_{TLM-small}$     & $\mathcal{C}_{TLM-small}$     & $\mathcal{C}_{TLM-small}$     & $\mathcal{C}_{TLM-small}$     & $\mathcal{C}_{TLM-small}$     & $\mathcal{C}_{TLM-small}$     &                      \\
\multicolumn{1}{l|}{}                                                                        & \multicolumn{1}{l|}{Training Data Size\tnote{2}}        & 0.22GB               & 0.04GB               & 0.1GB                & 0.18GB               & 0.3GB                & 0.32GB               & 0.13GB               & 0.16GB               &                      \\
\multicolumn{1}{l|}{}                                                                        & \multicolumn{1}{l|}{Training Steps}            & 5E4                  & 2E4                  & 1E5                  & 1E5                  & 1E5                  & 1E5                  & 1E5                  & 5E4                  &                      \\
\multicolumn{1}{l|}{}                                                                        & \multicolumn{1}{l|}{$\rho1$}        & 1                    & 99                   & 1                    & 19                   & 999                  & 999                  & 999                  & 3                    &                      \\
\multicolumn{1}{l|}{}                                                                        & \multicolumn{1}{l|}{$\rho2$}        & 100                  & 20                   & 100                  & 100                  & 100                  & 20                   & 20                   & 20                   &                      \\
\multicolumn{1}{l|}{}                                                                        & \multicolumn{1}{l|}{Batch Size}                & 256                  & 256                  & 256                  & 256                  & 256                  & 256                  & 256                  & 256                  &                      \\
\multicolumn{1}{l|}{}                                                                        & \multicolumn{1}{l|}{Sequence Length}           & 128                  & 128                  & 128                  & 512                  & 128                  & 128                  & 128                  & 128                  &                      \\ \cline{1-10}
\multicolumn{1}{l|}{\multirow{9}{*}{\begin{tabular}[c]{@{}l@{}}Medium\\ Scale\end{tabular}}} & \multicolumn{1}{l|}{$B_p$}      & 4                    & 1                    & 4                    & 4                    & 4                    & 4                    & 4                    & 4                    &                      \\
\multicolumn{1}{l|}{}                                                                        & \multicolumn{1}{l|}{$B_t$}          & 16                   & 1                    & 16                   & 16                   & 2                    & 4                    & 8                    & 16                   & \multicolumn{1}{c}{} \\
\multicolumn{1}{l|}{}                                                                        & \multicolumn{1}{l|}{Source Corpus\tnote{1}}             & $\mathcal{C}_{TLM-small}$     & $\mathcal{C}_{TLM-small}$     & $\mathcal{C}_{TLM-small}$     & $\mathcal{C}_{TLM-small}$     & $\mathcal{C}_{TLM-small}$     & $\mathcal{C}_{TLM-small}$     & $\mathcal{C}_{TLM-small}$     & $\mathcal{C}_{TLM-small}$     & \multicolumn{1}{c}{} \\
\multicolumn{1}{l|}{}                                                                        & \multicolumn{1}{l|}{Training Data Size\tnote{2}}        & 0.22GB               & 0.04GB               & 0.1GB                & 0.18GB               & 0.3GB                & 0.32GB               & 0.13GB               & 0.16GB               & \multicolumn{1}{c}{} \\
\multicolumn{1}{l|}{}                                                                        & \multicolumn{1}{l|}{Training Steps}            & 1.5E5                & 5E4                  & 1.5E5                & 1.5E5                & 1.5E5                & 1.5E5                & 1.5E5                & 1.5E5                & \multicolumn{1}{c}{} \\
\multicolumn{1}{l|}{}                                                                        & \multicolumn{1}{l|}{$\rho1$}        & 1                    & 99                   & 1                    & 19                   & 999                  & 999                  & 999                  & 3                    & \multicolumn{1}{c}{} \\
\multicolumn{1}{l|}{}                                                                        & \multicolumn{1}{l|}{$\rho2$}        & 100                  & 20                   & 100                  & 100                  & 100                  & 20                   & 20                   & 20                   & \multicolumn{1}{c}{} \\
\multicolumn{1}{l|}{}                                                                        & \multicolumn{1}{l|}{Batch Size}                & 256                  & 256                  & 256                  & 256                  & 256                  & 256                  & 256                  & 256                  & \multicolumn{1}{c}{} \\
\multicolumn{1}{l|}{}                                                                        & \multicolumn{1}{l|}{Sequence Length}           & 128                  & 128                  & 128                  & 512                  & 128                  & 128                  & 128                  & 128                  & \multicolumn{1}{c}{} \\ \cline{1-10}
\multicolumn{1}{l|}{\multirow{9}{*}{\begin{tabular}[c]{@{}l@{}}Large\\ Scale\end{tabular}}}  & \multicolumn{1}{l|}{$B_p$}      & 4                    & 1                    & 4                    & 4                    & 4                    & 4                    & 4                    & 4                    &                      \\
\multicolumn{1}{l|}{}                                                                        & \multicolumn{1}{l|}{$B_t$}          & 16                   & 1                    & 16                   & 16                   & 2                    & 4                    & 8                    & 16                   &                      \\
\multicolumn{1}{l|}{}                                                                        & \multicolumn{1}{l|}{Source Corpus\tnote{1}}             & $\mathcal{C}_{TLM-large}$     & $\mathcal{C}_{TLM-large}$     & $\mathcal{C}_{TLM-large}$     & $\mathcal{C}_{TLM-large}$     & $\mathcal{C}_{TLM-large}$     & $\mathcal{C}_{TLM-large}$     & $\mathcal{C}_{TLM-large}$     & $\mathcal{C}_{TLM-large}$     &                      \\
\multicolumn{1}{l|}{}                                                                        & \multicolumn{1}{l|}{Training Data Size\tnote{2}}        & 0.62GB               & 0.18GB               & 0.34GB               & 2.20GB               & 0.70GB               & 0.84GB               & 0.5GB                & 0.44GB               &                      \\
\multicolumn{1}{l|}{}                                                                        & \multicolumn{1}{l|}{Training Steps}            & 3E5                  & 1E5                  & 3E5                  & 3E5                  & 3E5                  & 3E5                  & 3E5                  & 3E5                  &                      \\
\multicolumn{1}{l|}{}                                                                        & \multicolumn{1}{l|}{$\rho1$}        & 3                    & 99                   & 1                    & 99                   & 999                  & 999                  & 999                  & 3                    &                      \\
\multicolumn{1}{l|}{}                                                                        & \multicolumn{1}{l|}{$\rho2$}        & 100                  & 100                  & 1000                 & 100                  & 20                   & 20                   & 100                  & 100                  &                      \\
\multicolumn{1}{l|}{}                                                                        & \multicolumn{1}{l|}{Batch Size}                & 256                  & 256                  & 256                  & 256                  & 256                  & 256                  & 256                  & 256                  &                      \\
\multicolumn{1}{l|}{}                                                                        & \multicolumn{1}{l|}{Sequence Length}           & 128                  & 128                  & 128                  & 512                  & 128                  & 128                  & 128                  & 128                  &                      \\ \hline
        
\end{tabular}

 }

\begin{tablenotes}
    \scriptsize
    \item{1} $\mathcal{C}_{TLM-small}$ and $\mathcal{C}_{TLM-large}$ are provided by TLM\cite{yao2022nlp}. 

    \item{2} ISS only uses a tiny subset of the source general corpus for training. We list the data size that are actually used for ISS training.
  \end{tablenotes}
 \end{threeparttable}
 \caption{Detailed hyper-parameters for ISS of different scales for each task}
 \label{DetailedSettingsScratch}
\end{table*}

\begin{table}[t]
 \begin{threeparttable}
\setcounter{table}{1}
\large
\centering
\renewcommand\arraystretch{1.5}
 \resizebox{1\columnwidth}{!}{
\begin{tabular}{lcccc}
\hline
\multicolumn{1}{l|}{\textbf{Hyper-Parameters}} & \textbf{RCT}         & \textbf{Chem.}       & \textbf{Acl.}        & \textbf{SciERC}      \\ \hline
\multicolumn{1}{l|}{$B_p$}      & 16                   & 8                    & 2                    & 4                    \\
\multicolumn{1}{l|}{$B_t$}          & 4                    & 4                    & 4                    & 4                    \\
\multicolumn{1}{l|}{Source Corpus\tnote{1}}             & $\mathcal{C}_\mathrm{S2ORC}$         & $\mathcal{C}_\mathrm{S2ORC}$         & $\mathcal{C}_\mathrm{S2ORC}$         & $\mathcal{C}_\mathrm{S2ORC}$         \\
\multicolumn{1}{l|}{Train Data Size\tnote{2}}           & 1.5G                 & 1.5G                 & 1.9G                 & 1.9G                 \\
\multicolumn{1}{l|}{Training Steps}            & 5E4                  & 5E4                  & 5E4                  & 5E4                  \\
\multicolumn{1}{l|}{Batch Size}                & 256                  & 256                  & 256                  & 256                  \\
\multicolumn{1}{l|}{Sequence Length}           & 128                  & 128                  & 128                  & 128                  \\ \hline
\end{tabular}
 }
\begin{tablenotes}
   
    \scriptsize
    \item{1} $\mathcal{C}_\mathrm{S2ORC}$ is provided by S2ORC\cite{lo-etal-2020-s2orc}.  
    \item{2} ISS only uses a tiny subset of the source general corpus for training. We list the data size that are actually used for ISS training.
  \end{tablenotes}
 \end{threeparttable}
 
 \caption{Detailed hyper-parameters for ISS in further pretraining}
 \label{table:furtherSettings}
\end{table}

\end{document}